\documentclass[11pt,a4paper]{article}
\usepackage[hyperref]{emnlp2020}
\usepackage{times}
\usepackage{latexsym}

\usepackage{microtype}

\aclfinalcopy 
\def\aclpaperid{1446} 

\newcommand{\short}{\textsc{Optimus}}
\newcommand{\longname}{\textbf{O}rganizing sentences via \textbf{P}re-\textbf{T}ra\textbf{i}ned \textbf{M}odeling of a \textbf{U}niversal \textbf{S}pace}

\usepackage{amsmath}
\usepackage{amsfonts}
\usepackage{amssymb}
\usepackage{wrapfig}
\usepackage{subcaption}
\usepackage{multirow}
 \usepackage{mathtools} 

\usepackage{verbatim}

\usepackage{anyfontsize}

\usepackage{color}
\usepackage{tikz}
\usetikzlibrary{arrows,shapes,snakes,automata,backgrounds,fit,petri}
\usepackage{adjustbox}

\newcommand{\RN}[1]{%
	\textup{\lowercase\expandafter{\it \romannumeral#1}}%
}

\usepackage{booktabs}

\usepackage{tcolorbox}

\makeatletter
\newcommand{\distas}[1]{\mathbin{\overset{#1}{\kern\z@\sim}}}%

\usepackage{enumitem}

\usepackage[lined,boxed,commentsnumbered,ruled,linesnumbered]{algorithm2e}

\newcommand{\ie}[0]{\emph{i.e., }}

\newcommand{\etc}[0]{\emph{etc.}}

\newcommand{\beq}{\vspace{0mm}\begin{equation}}
\newcommand{\eeq}{\vspace{0mm}\end{equation}}
\newcommand{\beqs}{\vspace{0mm}\begin{eqnarray}}
\newcommand{\eeqs}{\vspace{0mm}\end{eqnarray}}
\newcommand{\barr}{\begin{array}}
\newcommand{\earr}{\end{array}}

\newcommand{\Wmat}[0]{{{\bf W}}}

\newcommand{\bv}[0]{{\boldsymbol{b}}}
\newcommand{\cv}[0]{{\boldsymbol{c}}}

\newcommand{\hv}[0]{{\boldsymbol{h}}}

\newcommand{\xv}{\boldsymbol{x}}

\newcommand{\zv}{\boldsymbol{z}}

\newcommand{\thetav}{\boldsymbol{\theta}}

\newcommand{\phiv}{\boldsymbol{\phi}}

\newcommand{\R}{\mathbb{R}}
\newcommand{\E}{\mathbb{E}}

\newcommand{\Lcal}{\mathcal{L}}

\newcommand{\Fcal}{\mathcal{F}}

\ifx\assumption\undefined
\newtheorem{assumption}{Assumption}
\fi

\ifx\definition\undefined
\newtheorem{definition}{Definition}
\fi

\ifx\remark\undefined
\newtheorem{remark}{Remark}
\fi

\usepackage{color, colortbl}
\definecolor{Gray}{gray}{0.93}

\title{\textsc{Optimus}: Organizing Sentences via \\Pre-trained Modeling of a Latent Space}

\author{Chunyuan Li, Xiang Gao, Yuan Li, Baolin Peng, Xiujun Li, Yizhe Zhang, Jianfeng Gao 
\\
Microsoft Research, Redmond 
\\ {\small \texttt{\{chunyl, xiag, v-liyua, bapeng, xiul, yizzhang, jfgao\}@microsoft.com}} 
}

\date{}

\begin{document}
\maketitle

\begin{abstract}
When trained effectively, the Variational Autoencoder (VAE)~\cite{kingma2013auto,bowman2015generating} can be both a powerful generative model and an effective representation learning framework for natural language. 
In this paper, we propose the first large-scale language VAE model {\it {\short{}}} \footnote{\longname{}}. A universal latent embedding space for sentences is first pre-trained on large text corpus, and then fine-tuned for various language generation and understanding tasks. 
Compared with GPT-2, \short{} enables guided language generation from an abstract level using the latent vectors. Compared with BERT, \short{} can generalize better on low-resource language understanding tasks due to the smooth latent space structure. Extensive experimental results on a wide range of language tasks demonstrate the effectiveness of \short{}. It achieves new state-of-the-art on VAE language modeling benchmarks.
\end{abstract}

\section{Introduction}

Pre-trained language models (PLMs) have substantially advanced the state-of-the-art across a variety of natural language processing (NLP) tasks~\cite{peters2018deep,devlin2019bert,yang2019xlnet,radford2019language,liu2019roberta,keskar2019ctrl,shoeybi2019megatron}. PLMs are often trained to predict words based on their context on massive text data, and the learned models can be fine-tuned to adapt to various downstream tasks.

PLMs can generally play two different roles: 
$(\RN{1})$ 
a generic {\em encoder} such as BERT~\cite{devlin2019bert} to provide contextualized representations for language understanding tasks, and 
$(\RN{2})$ 
a powerful {\em decoder} such as GPT-2~\cite{radford2019language} to generate text sequences in an auto-regressive manner. In a bid to combine language understanding and generation tasks in one unified framework, several model variants have been proposed, including UniLM~\cite{dong2019unified}, BART~\cite{lewis2019bart}, and T5~\cite{raffel2019exploring}.  Although significant performance improvement has been reported on a wide range of NLP tasks, these models lack of explicit modeling of structures in a compact latent space, rendering it difficult to control language generation/representation from an abstract level.

Variational Autoencoders (VAEs)~\cite{kingma2013auto,rezende2014stochastic} provide a tractable method to train latent-variable generative models. In NLP,  latent variables may assume the role of 
higher-level sentence representations, which govern a lower-level word-by-word generation process, thus facilitating controlled text generation~\cite{bowman2015generating,hu2017toward}. By representing sentences in a low-dimensional latent space, VAEs allow easy manipulation of sentences using the corresponding {\em compact} vector representations, such as feature regularization specified by prior distributions, and guided sentence generation with interpretable vector operators. 
Despite the attractive theoretical strengths, the current language VAEs are often built with shallow network architectures, such as two-layer LSTMs~\cite{hochreiter1997long}. This limits the model's capacity and leads to sub-optimal performance.  

In this paper, we propose \short{}, the first large-scale pre-trained deep latent variable models for natural language. \short{} is pre-trained using the sentence-level (variational) auto-encoder objectives on large text corpus. This leads to a universal latent space to organize sentences (hence named \short{}). 
\short{} enjoys several favorable properties:
$(\RN{1})$ 
It combines the strengths of VAE, BERT and GPT, and supports both natural language understanding and generation tasks. 
$(\RN{2})$ 
Comparing to BERT, \short{} learns a more structured semantic space due to the use of the prior distribution in training. As a result, the language representations learned by \short{} are more universal / general in that they can be more easily adapted to a new domain/task.
$(\RN{3})$ 
Different from GPT-2, which generates human-like text but may lack effective means of controlling its high-level semantics (such as tense, topics, sentiment), \short{} can be easily deployed for guided text generation.      
The effectiveness of \short{} has been demonstrated with extensive experiments on language modeling, dialog response generation, text style transfer and low-resource language understanding. It achieves lower perplexity than GPT-2 on standard benchmarks, produces strong performance on guided text generation, and improves BERT on feature-based language understanding tasks.  
The code and pre-trained models are released on Github\footnote{\url{https://github.com/ChunyuanLI/Optimus}}.

Along the way to build the first big VAE language model, there are several technical contributions/implications that are novel: $(\RN{1})$  Latent vector injection: this work demonstrates two schemes to discuss how to effectively inject conditioning vectors into GPT-2 without re-training it. 
$(\RN{2})$  The design idea to combine BERT/GPT-2 serves as a practical recipe to inspire people to integrate and reuse existing PLMs for larger and complex models. 
$(\RN{3})$  Pre-training on massive datasets itself is an effective approach to reduce KL vanishing, as demonstrated by the state of-the-art performance on four VAE language modeling datasets. $(\RN{4})$  The proof of VAE objective from the lens of IB, showing that VAE is a principled approach to balance the compactness and usability of learned representations. 
$(\RN{5})$  Improved performance on several language tasks shows the importance and necessity of pre-training a latent space.

\section{Related Work}
\vspace{-2mm}
\paragraph{Difference with prior PLMs.} Large-scale Transformer-based PLMs have recently achieved state-of-the-art performance on various natural language understanding and generation tasks~\cite{devlin2019bert,yang2019xlnet,radford2019language,liu2019roberta,keskar2019ctrl}.
Prior to Transformer-based PLMs, non-generative methods have seen some early success in pre-training sequence models for supervised downstream tasks
including standard sequence auto-encoders~\cite{dai2015semi,li2015hierarchical}, skip-thought models~\cite{kiros2015skip} and paragraph vector models~\cite{le2014distributed} \etc~
However, all of these models do not generally learn a smooth, interpretable feature space for sentence encoding, or generating novel sentences. In this work, we aim to fill the gap to learn such a universal latent space in the field of Transformer-based PLMs.

\paragraph{Latent variable language modeling.} 
Language VAEs have inspired new applications in NLP, via exploiting many interesting
properties of the model's latent space~\cite{bowman2015generating,kim2018tutorial}. 
Its modeling capacity and empirical performance is somewhat limited, partially due to the KL vanishing issue described in Section~\ref{sec:learning_procedure}. Several attempts have been made to alleviate this issue, including different KL annealing/thresholding schemes~\cite{bowman2015generating,fu2019cyclical,higgins2017beta,li2019surprisingly}, decoder architectures~\cite{yang2017improved,dieng2018avoiding}, auxiliary loss~\cite{zhao2017learning}, semi-amortized inference~\cite{kim2018semi}, aggressive encoder training schedule~\cite{he2019lagging}, batch normalized inference~\cite{zhu2020batch} and flexible posterior~\cite{fang2019implicit}. \citet{subramanian2018towards} have shown some promise that general encoder can benefit language generation. Transformers~\cite{vaswani2017attention} are recently considered in VAEs for  classification~\cite{gururangan2019variational} and storytelling~\cite{wang2019t}. Pre-training VAEs has been recently considered in conditional text generation to amortize the training of decoders and to allow easy adaptation in new generation tasks~\cite{duan2019pre}.

All these efforts utilize simple LSTM~\cite{hochreiter1997long} and shallow Transformer~\cite{vaswani2017attention} architectures, thus with limited capacity. Our paper is the first big VAE model at the same scale of recent PLMs such as BERT and GPT-2. More importantly, we show that pre-training a meaningful latent space on a large text corpus can largely reduce the KL vanishing issue, and lead to new state-of-the-art performance.

\section{Background on NLMs \& GPT-2}
\vspace{-2mm}

To generate a text sequence of length $T$, $\xv =[x_1, \cdots, x_T]$, neural language models (NLM)~\cite{mikolov2010recurrent} generate every token $x_t$ conditioned on the previous word tokens:
\begin{align}
\vspace{-2mm}
\label{eq_lm_generator}
p(\xv) = \prod_{t=1}^{T} p_{\thetav}(x_t| x_{<t}), 
\vspace{-2mm}
\end{align}
where $x_{<t}$ indicates all tokens before $t$, and $\thetav$ is the model parameter. 
In NLMs, each one-step-ahead conditional in~\eqref{eq_lm_generator} is modeled by an expressive family of neural networks, and is typically trained via maximum likelihood estimate (MLE).
Perhaps the most well-known NLM instance is GPT-2~\cite{radford2019language}, which employs Transformers~\cite{vaswani2017attention} for each conditional, and $\thetav$ is learned on a huge amount of OpenWeb text corpus. GPT-2 has shown surprisingly realistic text generation results, and low perplexity on several benchmarks. GPT-3~\cite{brown2020language} was recently proposed to further scale up NLMs to 175 billion parameters, showing impressive results on few-shot learning on multiple language tasks.

However, the only source of variation in NLMs, GPT2 and GPT3 is modeled in the conditionals at every step: the text generation process only depends on previous word tokens, and there is limited capacity for the generation to be guided by the higher-level structures that are likely presented in natural language, such as tense, topics or sentiment.

\vspace{-0mm}
\section{Pre-trained Latent Space Modeling}
\label{sec:pre_train}
\vspace{-0mm}
%

\subsection{Pre-training Objectives}
\label{sec:pre_train_objective}
To facilitate high-level guidance in sentence generation, 
\short{} organizes sentences in a universal latent (or semantic) space, via pre-training on large text corpora. Each sample in this space can be interpreted as outlines of the corresponding sentences, guiding the language generation process performed in the symbolic space~\cite{subramanian2018towards}. This naturally fits within the learning paradigm of latent variable models such as VAEs~\cite{kingma2013auto,bowman2015generating}, where the latent representations capture the high-level semantics/patterns.  
It consists of two parts, generation and inference, enabling a bidirectional mapping between the latent space and symbolic space.

%


\paragraph{Generation} The {\it generative model} ({\it decoder}) draws a latent vector $\zv$ from the continuous latent space with prior
$p(\zv)$, and generates the text sequence $\xv$ from a conditional distribution $p_{\thetav}(\xv|\zv)$; $p(\zv)$ is typically assumed a multivariate Gaussian, and $\thetav$ represents the neural network parameters. The following auto-regressive decoding process is usually used: 
\begin{align}
\vspace{-2mm}
p_{\thetav}(\xv |\zv) = \prod_{t=1}^{T} p_{\thetav}(x_t| x_{<t}, \zv).
\label{eq_vae_decoder}
\vspace{-4mm}
\end{align}
Intuitively, VAE provides a ``hierachical'' generation procedure: $\zv \sim p(\zv)$ determines the high-level semantics, followed by \eqref{eq_vae_decoder} to produce the output sentences with low-level syntactic and lexical details. This contrasts with~\eqref{eq_lm_generator} in the explicit dependency on $\zv$. 

\paragraph{Inference} Similar to GPT-2, parameters $\thetav$ are typically learned by maximizing the marginal log likelihood $\log p_{\thetav}(\xv) =
\log \int p(\zv) p_{\thetav}(\xv |\zv) \mbox{d} \zv$.   
However, this marginal term is intractable to compute for many decoder choices. Thus, variational inference is considered, and the true posterior $p_{\thetav} (\zv | \xv) \propto p_{\thetav} (\xv | \zv) p(\zv) $ is approximated via
the variational distribution $q_{\phiv}(\zv | \xv)$ is (often known as the {\it inference model} or {\it encoder}), implemented via a $\phiv$-parameterized neural network. 
It yields the {\it evidence lower bound objective} (ELBO):
\begin{align}
\label{eq_vae_elbo}
& \log p_{\thetav}(\xv)  \ge \Lcal_{ \text{ELBO} } = \\  
& \E_{q_{\phiv}(\zv | \xv)} \big[ \log p_{\thetav}(\xv | \zv) \big]
-\mbox{KL} (q_{\phiv}(\zv | \xv) || p(\zv) ) 
\nonumber
\end{align}

Typically, $q_{\phiv} (\zv | \xv)$ is modeled as a Gaussian distribution, and the re-parametrization trick is used for efficient learning~\cite{kingma2013auto}.

\paragraph{A Taxonomy of Autoencoders}
There is an alternative interpretation of the ELBO: the VAE objective can be viewed as a regularized version of the autoencoder (AE)~\cite{goodfellow2016deep}. 
It is thus natural to extend the negative of $ \Lcal_{ \text{ELBO} }$ in \eqref{eq_vae_elbo} by introducing a hyper-parameter $\beta$ to control the strength of regularization:
\begin{align}
\Lcal_{\beta}  
 & = \Lcal_{E} +\beta \Lcal_{R},~~\text{with} 
 \label{eq_elbo_beta} 
\\
\Lcal_{E} & = -\E_{q_{\phiv}(\zv | \xv)} \big[ \log p_{\thetav}(\xv | \zv) \big]  
\label{eq_rec} 
\\
\Lcal_{R} & = \mbox{KL} (q_{\phiv}(\zv | \xv) || p(\zv) ) 
\label{eq_kl}
\end{align}
where $\Lcal_{E}$ is the reconstruction error (or negative log-likelihood (NLL)), and $\Lcal_{R}$ is a KL regularizer.
The cost function $\Lcal_{\beta}$ provides a unified perspective for understanding various autoencoder variants and training methods. We consider two types of latent space with the following objectives:

\begin{itemize}
    \item {\bf AE}. Only $\Lcal_{E}$ is considered ($\beta = 0$), while the Gaussian sampling in $q_{\phiv}(\zv | \xv)$ remains. In other words, the regularization is removed, and a point-estimate is likely to be learned to represent the text sequence's latent feature. 
    Note our reconstruction is on sentence-level, while other PLMs~\cite{devlin2019bert,yang2019xlnet} employ masked LM loss, performing token-level reconstruction. \vspace{-2mm}
    \item {\bf VAE}. The full VAE objective is considered ($\beta > 0$). It tends to learn a smooth latent space due to $\Lcal_{R}$. 
    \vspace{-0mm}
\end{itemize}
%

\paragraph{Information Bottleneck Principle} 
From an information theory perspective, {\em  information bottleneck} (IB) provides a principled approach to find the trade-off between {\em predictive power} and {\em complexity (compactness)} when summarizing observed data in learned representations. We show that our \short{} pre-training objectives effectively practice the IB principle as follows.

The objective in~\eqref{eq_elbo_beta} shows the $\beta$-VAE loss for one single sentence $\xv$. The training objective over the dataset $q(\xv)$ can be written as:
\begin{align} \label{eq_agg_vae_analysis}
\Fcal_{\beta} = -\Fcal_E + \beta \Fcal_R 
\vspace{-2mm}
\end{align}
where $\Fcal_E = E_{q(\xv), \zv \sim q(\zv|\xv)} [\log p(\Tilde{\xv} | \zv ) ]$ is the aggregated reconstruction term ($\Tilde{\xv}$ is the reconstruction target), and $\Fcal_R=\E_{q(\xv)}[ \mbox{KL} (q(\zv | \xv) || p(\zv) ) ] $ is the aggregated KL term. With the detailed proof shown in Section~\ref{supp_sec:ib_proof} of Appendix, we see that $\Fcal_\beta$ is an upper bound of IB:
\begin{align} \label{eq_vae_bound_bi}
\Fcal_{\beta} \ge - I_q(\zv, \Tilde{\xv} ) +  \beta I_q(\zv, \xv) = \Lcal_{\text{IB}}, 
\end{align}
where $\Lcal_{\text{IB}}$ is the Lagrange relaxation form of IB presented by~\citet{tishby2000information}, $I_q(\cdot,\cdot)$ is the mutual information (MI) measured by probability $q$. The goal of IB is to maximize the predictive power of $\zv$ on target $\Tilde{\xv}$, subject to the constraint on the amount of information about original $\xv$ that $\zv$ carries.
When $\beta=0$, we have the AE variant of our \short{}, the model fully focuses on maximizing the MI to recover sentences from the latent space. As $\beta$ increases, the model gradually transits towards fitting the aggregated latent distribution $q(\zv)=\int_{\xv} q(\zv | \xv) q(\xv) d \xv $ to the given prior $p(\zv)$, leading the VAE variant of our \short{}.

\begin{figure}[t!]
	\vspace{-0mm}\centering
	\begin{tabular}{c}
		\hspace{-3mm}
		\includegraphics[height=1.7cm]{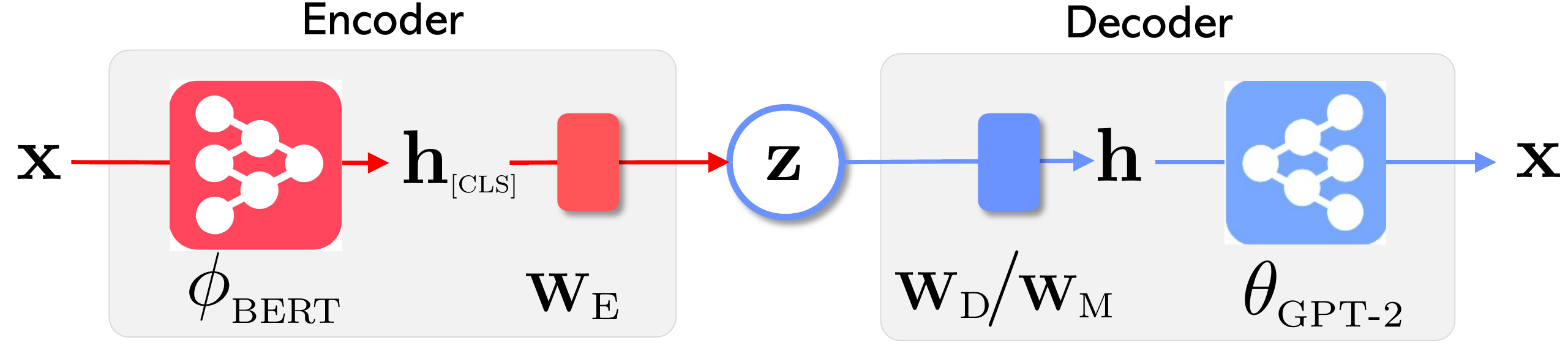} 
	\end{tabular}
	\vspace{-0mm}
	\caption{Illustration of \short{} architecture.
	 }
	\vspace{-0mm}
	\label{fig:optimus_schemes}
\end{figure}

\subsection{Model Architectures}
\vspace{-0mm}
The model architecture of \short{} is composed of multi-layer Transformer-based encoder and decoder, based on the original implementation described in~\cite{vaswani2017attention}. The overall architecture is illustrated in Figure~\ref{fig:optimus_schemes}.  
To leverage the expressiveness power of existing PLMs, we initialize our encoder and decoder with weights of BERT $\phiv_{\text{BERT}}$  and GPT-2 $\thetav_{\text{GPT-2}}$, respectively. This procedure is seamless, as all of these models are trained in a self-supervised/unsupervised manner.

We denote the number of layers (\ie Transformer blocks) as $L$, the hidden size as $H$, and the number of self-attention heads as $A$. Specifically, we consider BERT$_{\text{BASE}}$ (L=12, H=768, A=12, Total Parameters=110M) and GPT-2 (L=12, H=768, A=12, Total Parameters=117M).
We hope that our approach can provide a practical recipe to inspire future work to integrate larger pre-trained encoder and decoder for higher performance models.

\paragraph{Connecting BERT \& GPT-2} Two technical questions remain, when pre-training \short{} from BERT \& GPT-2: 
$(\RN{1})$ How to represent sentences, since the two PLMs employ different tokenization schemes?
$(\RN{2})$ How to adapt a pre-trained GPT-2 to arbitrary conditional input without re-training the model again? Controllable GPT-2 models have been studied in~\cite{keskar2019ctrl,zellers2019defending,peng2020soloist,peng2020few} when prescribed control codes/tokens are provided, but it is still unknown how to ground GPT-2 to arbitrary conditional inputs.

\paragraph{Tokenization}
In BERT, WordPiece Embeddings (WPE) is used for tokenization (vocabulary size is 28996 for the cased version). In GPT-2, the modified Byte Pair Encoding (BPE)~\cite{radford2019language} is used for tokenization (vocabulary size is 50260). A given token is represented as $\hv_{\texttt{Emb}}$, by summing the corresponding token, position and segment embeddings~\footnote{\short{} does not require segment embeddings, but we remain it due to BERT initialization.}.
For a sentence, we present it in both types of tokenization: the input of encoder is WPE, and the output of decoder is BPE to compute the reconstruction loss.

\begin{figure}[t!]
	\vspace{-0mm}\centering
	\begin{tabular}{c c}
		\hspace{-4mm}
		\includegraphics[height=2.5cm]{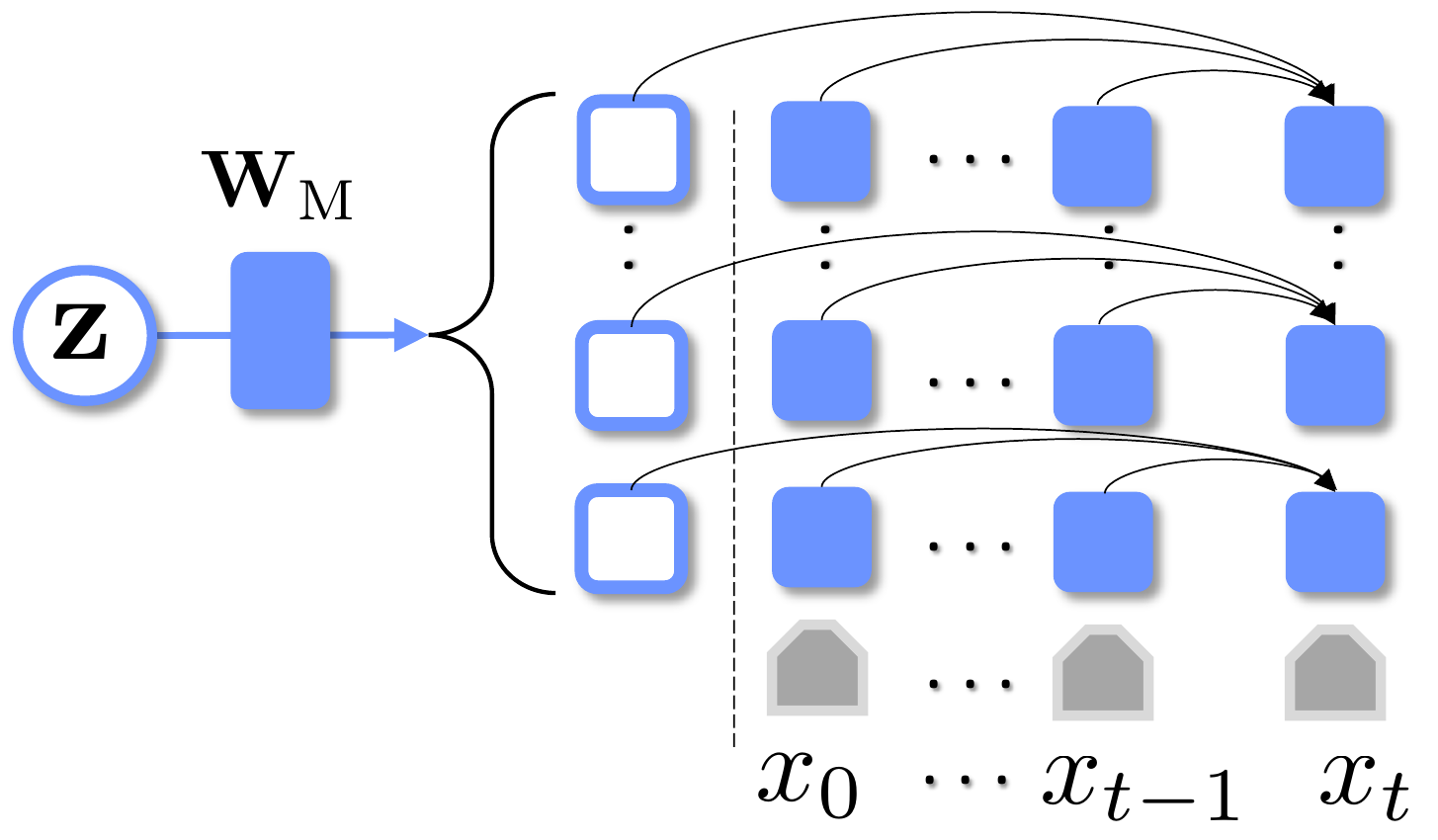}  & 
		\hspace{-3mm}
		\includegraphics[height=2.5cm]{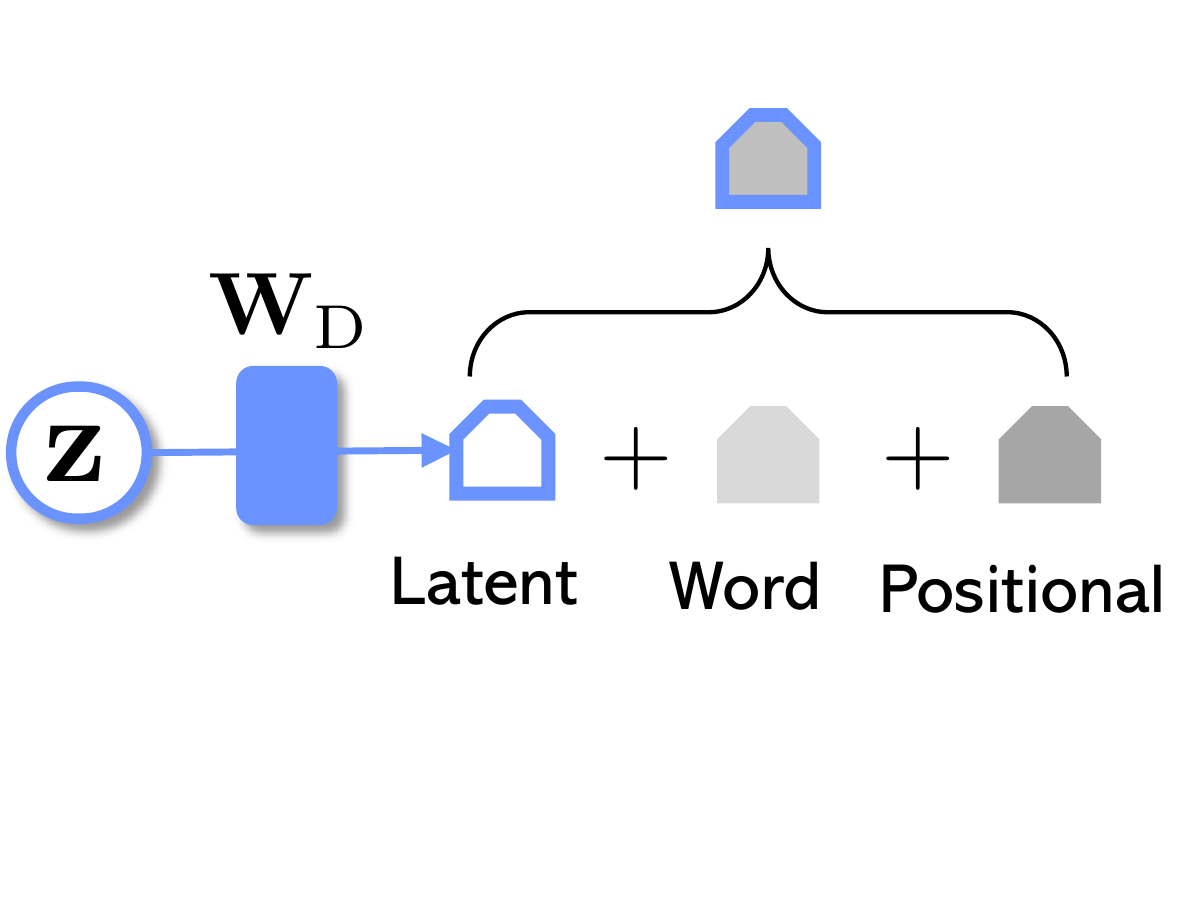} \\
		(a) Memory \vspace{0mm} & 
		(b) Embedding \hspace{-0mm} \\ 
	\end{tabular}
	\vspace{-0mm}
	\caption{Illustration of two schemes to inject latent vector. (a) Memory: $x_t$ attends both $x_{<t}$ and $\hv_{\texttt{Mem}}$; (b) Embedding: latent embedding is added into old embeddings to construct new token embedding $\hv_{\texttt{Emb}}^{\prime}$.
	 }
	\vspace{-0mm}
	\label{fig:latent_inject_schemes}
\end{figure}

\paragraph{Latent Vector Injection}
Similar to BERT, the first token of every sentence is always a special classification token (\texttt{[CLS]}). The last-layer hidden state $\hv_{\texttt{[CLS]}} \in \R^{H}$ corresponding to this token is used as the sentence-level representation. It further constructs the latent representation $\zv = \Wmat_{\text{E}} \hv_{\texttt{[CLS]}}$, where $\zv \in \R^{P}$ is a $P$-dimensional vector and $\Wmat_{\text{E}} \in \R^{P\times H}$ is the weight matrix. To facilitate $\zv$ in GPT-2 decoding without re-training the weights, we consider two schemes, illustrated in Figure~\ref{fig:latent_inject_schemes}:
\vspace{-2mm}
\begin{itemize}
    \item \texttt{Memory}:~$\zv$ plays the role of an additional {\em memory} vector $\hv_{\texttt{Mem}}$ for GPT2 to attend. Specifically, $\hv_{\texttt{Mem}} = \Wmat_{\text{M}} \zv$, where $\Wmat_{\text{M}} \in \R^{LH \times P}$ is the weight matrix. $\hv_{\texttt{Mem}} \in \R^{LH}$ is separated into $L$ vectors of length $H$, each of which is attended by GPT-2 in one layer. \vspace{-2mm}
    \item \texttt{Embedding}:~$\zv$ is added on the original {\em embedding} layer, and directly used in every decoding step. The new embedding representation is $\hv_{\texttt{Emb}}^{\prime} = \hv_{\texttt{Emb}} + \Wmat_{\text{D}}\zv$, where $\Wmat_{\text{D}} \in \R^{H \times P}$.
\end{itemize}
We study their empirical performance in Section \ref{supp_sec:inject} of Appendix, and observe that \texttt{Memory} is significantly more effective than \texttt{Embedding}, and the integration of both schemes yields slightly better results. We hypothesize that the reason why \texttt{Memory} is superior is because it allows the decoder to attend the latent information at every layer of
the network directly, while the \texttt{Embedding} method only allows the decoder to see the latent
information at the input and output layer.
In our experiments, we use the integration scheme by default.
In summary, the encoder parameters $\phiv = \{\phiv_{\text{BERT}}, \Wmat_{\text{E}} \}$, and decoder parameters $\thetav = \{\thetav_{\text{GPT-2}}, \Wmat_{\text{M}},  \Wmat_{\text{D}} \}$.

\subsection{Learning Procedures}
\label{sec:learning_procedure}

We train the model parameters $\{\phiv, \thetav\}$ using two objectives: AE and VAE, discussed in Section~\ref{sec:pre_train_objective}. Pre-training AE using ~\eqref{eq_rec}  is straightforward. However, pre-training VAE can be challenging due to the notorious {\em KL vanishing} issue~\cite{bowman2015generating}, where 
$(\RN{1})$ 
an encoder that produces posteriors almost identical to the Gaussian prior for all sentences (rather than a more interesting posterior); and 
$(\RN{2})$ 
a decoder that completely ignores $\zv$ in~\eqref{eq_vae_decoder}, and a learned model that reduces to a simpler NLM.

To reduce this issue, we follow the intuition that if the encoder is providing useful information from the beginning of decoder training, the decoder is more likely to make use of $\zv$~\cite{fu2019cyclical,he2019lagging}. Specifically, we use the cyclical schedule to anneal $\beta$ for 10 periods~\cite{fu2019cyclical}.
Within one period, there are three consecutive stages: Training AE ($\beta=0$) for 0.5 proportion, annealing $\beta$ from 0 to 1 for 0.25 proportion, and fixing $\beta=1$ for  0.25 proportion. When $\beta>0$, we use the KL thresholding scheme~\cite{li2019surprisingly,kingma2016improved}, and replace  the KL term $\Lcal_{R}$ in~\eqref{eq_kl} with a hinge loss term that maxes each component of the original KL with a constant $\lambda$:  
\begin{align}
\Lcal_{R}^{\prime} = \sum_{i}
 \max [\lambda, \mbox{KL}(q_{\phiv}(z_i | \xv) || p(z_i) )] 
\label{eq_kl_thresholding} 
\end{align}
Here, $z_i$ denotes the $i$th dimension of $\zv$. Using
the thresholding objective causes learning to give up driving down KL for dimensions of $\zv$ that are already beneath the target compression rate.


\paragraph{Pre-training data} The pre-training procedure largely follows the existing literature on language model pre-training. We use English Wikipedia to pre-train our AE and VAE objectives. As our main interest is to model sentences (rather than text sequences of a fixed length), we pre-process Wikipedia with maximum sentences length 64. It leads to 1990K sentences, which accounts 96.45\% Wikipedia sentences used in BERT. More data pre-processing details are in Section~\ref{supp_sec:wiki} of Appendix.






\begin{table*}[t!]
\centering
\includegraphics[width=1.95\columnwidth]{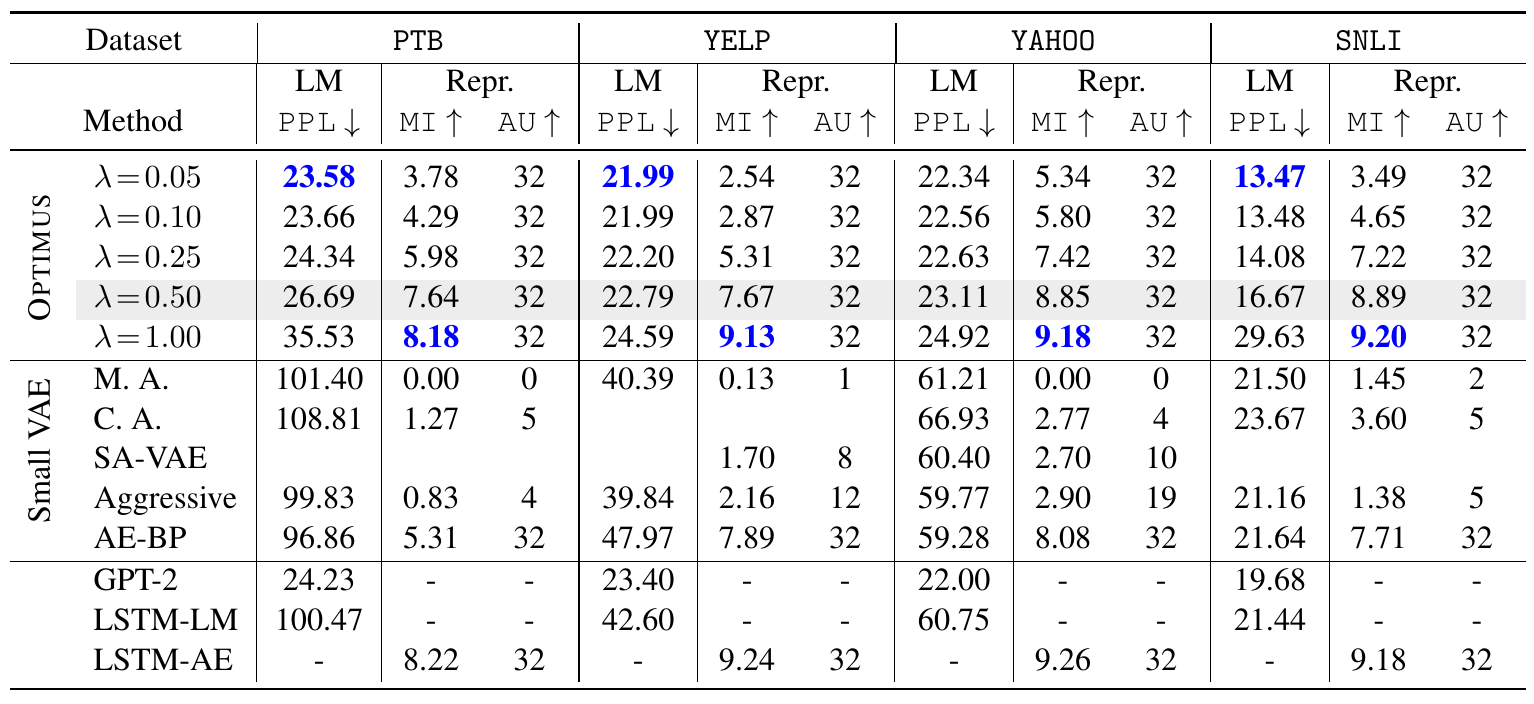}
\vspace{-0mm}
  \caption{Comparison on language modeling tasks on four datasets. ``Small VAEs'' indicate all previous language VAEs, which are built with two-layer LSTMs. All results for Small VAEs, LSTM-LM, LSTM-AE are quoted from literature, and GPT-2 results are produced by us. 
  Best values are in blue. $\lambda\!=\!0.50$ is a good trade-off to achieve the best values on all metrics compared with small VAEs. ``-'' indicates the models are improper to report these values; Empty cells indicate the results were not reported in the literature.}
  \label{tab:compare_sota_lm}
\vspace{-0mm}
\end{table*}

\section{Experimental Results}
We consider to apply the pre-trained \short{} models to three types of downstream tasks: 
$(\RN{1})$ language modeling, where \short{} is compared with SoTA VAE methods and GPT-2.
$(\RN{2})$ Guided language generation, where \short{} shows its unique advantage in producing controllable sentences in contrast to GPT-2.  
$(\RN{3})$ Low-resource language understanding, where the learned structured latent features can be used for fast adaptation in new tasks.

\vspace{-0mm}
\subsection{Language Modeling}
Fine-tuning LM on new datasets is straightforward. We load the pre-trained \short{}, and update the model with one additional $\beta$ scheduling cycle for one epoch. The semantic latent vectors are first pre-trained off-the-shelf, and then easily leveraged to train the decoder on downstream datasets. From this perspective, our pre-training can be viewed as an effective approach to reduce KL vanishing. 

We consider four datasets: the Penn Treebank ($\mathtt{PTB}$) \cite{marcus1993building}, $\mathtt{SNLI}$~\cite{bowman2015large}, $\mathtt{Yahoo}$, and  $\mathtt{Yelp}$  corpora~\cite{yang2017improved, he2019lagging}.
\paragraph{Metrics} There are two types of metrics to evaluate language VAEs. 
$(\RN{1})$  Generation capability: we use {\em perplexity} (PPL). Note that NLM and GPT-2 has exactly PPL, while VAEs does not. Following~\cite{he2019lagging}, we use the importance weighted bound in~\cite{burda2015importance} to approximate $\log p(\xv)$, and report PPL.
$(\RN{2})$ Representation learning capability: Active units (AU) of $\zv$ and its Mutual Information (MI) with $\xv$. 
We report the full results with ELBO, KL and Reconstruction in Appendix, but note that higher ELBO does not necessarily yield better language modeling. 

\paragraph{Baseline Methods}
$(\RN{1})$ {\em GPT-2}. A large-scale LM trained on OpoenWebText~\cite{radford2019language}. We load the pre-trained GPT-2 weights, and refine the model for 1 epoch on the new datasets.
$(\RN{2})$ {\em Annealing}. $\beta$ is gradually annealed from 0 to 1. This annealing procedure can be used once (M.A.)~\cite{bowman2015generating} or multiple times (C.A.)~\cite{fu2019cyclical}.
$(\RN{3})$ {\em Aggressive Training}~\cite{he2019lagging}. Training the encoder multiple times per decoder update.
$(\RN{4})$ {\em AE-FB}~\cite{li2019surprisingly}. Training AE, and then VAE using the KL thresholding in~\eqref{eq_kl_thresholding}, the results on $\lambda\!=\!0.50$ are reported as a good trade-off.

The results are shown in Table~\ref{tab:compare_sota_lm}. Various $\lambda$ values are used, we observe a trade-off between language modeling and representation learning, controlled by $\lambda$. Compared with existing VAE methods, \short{} achieve significantly lower perplexity, and higher MI/AU. This indicates that our pre-training method is an effective approach to reduce KL vanishing issue and training VAEs, especially given the fact that we only fine-tune on these datasets for one epoch. \short{} achieves lower perplexity compared with GPT-2 on three out of four datasets. Intuitively, this is because the model can leverage the prior language knowledge encoded in $\zv$. This gap is larger, when the sentences in the dataset exhibit common regularities, such as $\mathtt{SNLI}$, where the prior plays a more important/effective role in this scenario. 
Though the form of our model is simple, \short{} shows stronger empirical performance than sophisticated models that are particularly designed for long-text, such as hVAE in~\cite{shen2019towards}. For example, the KL and PPL of
\short{} (15.09 and 22.79) are much better than hVAE (6.8 and 45.8) on Yelp dataset. This verifies the importance of pre-training a latent space.
The full experimental results are shown in Table~\ref{tab:compare_sota_penn}, \ref{tab:compare_sota_yelp}, \ref{tab:compare_sota_yahoo} and \ref{tab:compare_sota_snili} of Appendix.

\begin{table*}[t!]\centering
\begin{minipage}{16cm}\vspace{0mm}    \centering
\begin{tcolorbox} 
\vspace{-2mm}
\begin{adjustbox}{scale=0.8,tabular=p{9cm} p{10cm},center}
{\bf Source} $\xv_A$ & {\bf Target}  $\xv_B$
\tabularnewline
a girl makes a silly face & two soccer players are playing soccer 
\tabularnewline \midrule
{\bf Input} $\xv_C$  & {\bf Output} $\xv_D$
\tabularnewline
$\bullet$ a girl poses for a picture &
$\bullet$ \textcolor{blue}{ two soccer players are at a soccer game.} 
\tabularnewline
$\bullet$  a girl in a blue shirt is taking pictures of a microscope & 
$\bullet$  \textcolor{blue}{ two football players in blue uniforms are at a field hockey game} \tabularnewline
$\bullet$ a woman with a red scarf looks at the stars & 
$\bullet$ \textcolor{blue}{two men in white uniforms are field hockey players}
\tabularnewline
$\bullet$ a boy is taking a bath & 
$\bullet$ \textcolor{blue}{ two baseball players are at the baseball diamond} \tabularnewline
$\bullet$ a little boy is eating a bowl of soup & 
$\bullet$ \textcolor{blue}{ two men are in baseball practice} 
\end{adjustbox}
\end{tcolorbox}
\vspace{-2mm}
\caption{Sentence transfer via arithmetic  $\zv_D= \zv_B - \zv_A + \zv_C$. The output sentences are in blue.} 
	\label{table:transfer_example_main}
\end{minipage}
\vspace{-0mm}
\end{table*}

\begin{table}[t!] 
    \begin{tcolorbox}[fontupper=\footnotesize, fontlower=\footnotesize]
      \hspace{-5mm}
    \begin{tabular}{l l}
$0.0$ & children are looking for the water to be clear. \tabularnewline
$0.1$ & \textcolor{blue}{children are looking for the water.} \tabularnewline
$0.2$ & \textcolor{blue}{children are looking at the water.} \tabularnewline
$0.3$ & \textcolor{blue}{the children are looking at a large group of people.} \tabularnewline
$0.4$ & \textcolor{blue}{the children are watching a group of people.} \tabularnewline
$0.5$ & \textcolor{blue}{the people are watching a group of ducks.} \tabularnewline
$0.6$ & \textcolor{blue}{the people are playing soccer in the field.}\tabularnewline
$0.7$ & \textcolor{blue}{there are people playing a sport.} \tabularnewline
$0.8$ & \textcolor{blue}{there are people playing a soccer game.} \tabularnewline
$0.9$ & \textcolor{blue}{there are two people playing soccer.} \tabularnewline
$1.0$ & there are two people playing soccer. 
	\end{tabular}
	\end{tcolorbox}
	\vspace{-2mm}
	\caption{Interpolating latent space $\zv_{\tau}=\zv_{1}\cdot(1-\tau)+\zv_{2}\cdot \tau$. Each row shows $\tau$, and the generated sentence (in blue) conditioned on $\zv_{\tau}$.}
	\label{table:interpolation_example_main}
	\vspace{-0mm}
\end{table}

\vspace{-0mm}
\subsection{Guided Language Generation}
\vspace{-0mm}

Different from the traditional NLMs or GPT-2, VAEs learns bidirectional mappings between the latent and symbolic space. It enables high-level sentence editing as arithmetic latent vector operations, and thus allows guided language generation.
The reason that Optimus supports arithmetic operations are two-fold: (1) Pre-training on large datasets
with large networks allows all sentences to be densely and faithfully represented in the latent
space. (2) The continuity property of neural nets and KL regularization of VAE encourage latent
vectors with similar semantics are smoothly organized together.

This is demonstrated with two simple schemes to manipulate pre-trained latent spaces: sentence transfer and interpolation, with results in Table~\ref{table:transfer_example_main} and Table~\ref{table:interpolation_example_main}, respectively. 
Details and more results are shown in Appendix.
They showcase that \short{} enables new ways that one can play with language generation using pre-trained models, compared with GPT-2 that can only fulfill text sequences with given prompts. A website demo\footnote{\url{http://aka.ms/optimus}} is released to the public to interact with the model, exhibiting the power of latent-vector-based controllable text generation. 
We demonstrate more sophisticated ways to manipulate pre-trained latent spaces in three real applications as follows.

\paragraph{Dialog response generation}
The open-domain dialog response generation task is considered: generating responses $\xv$ given a dialog history $\cv$. Following~\cite{gao2019jointly}, we embed the history and response in a joint latent space as $\zv_{\text{S2S}}$ and $ \zv_{\text{AE}}$, respectively. A fusion regularization is used to match the responses to the context.
We consider $\mathtt{Dailydialog}$ \cite{li2017dailydialog} used in~\cite{gu2018dialogwae}, which has 13,118 daily conversations. Each utterance is processed as the response of previous 10 context utterances from both speakers. The baseline methods are described in Appendix. We measure the performance using Bleu~\cite{chen2014systematic}, and compute the precision, recall and F1 in Table~\ref{table:dialogue}. \short{} shows higher Bleu scores than all existing baselines.

\begin{table}[t!]
\begin{centering}
\includegraphics[width=0.95\columnwidth]{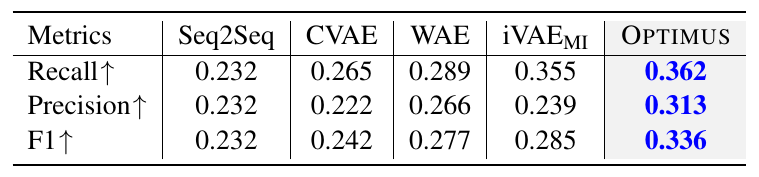}
\end{centering}
\vspace{-2mm}
\caption{Dialog response generation on  $\mathtt{DailyDialog}$ dataset. All numbers are  from~\cite{gu2018dialogwae} except that iVAE$_{\text{MI}}$ is
from~\cite{fang2019implicit}.}
\label{table:dialogue}
\vspace{-0mm}
\end{table}

\begin{table}[t!]
\begin{centering}
\includegraphics[width=0.95\columnwidth]{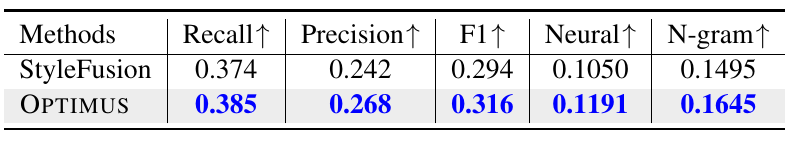}
\end{centering}
\vspace{-2mm}
\caption{Stylized response generation.}
\label{table:dialogue_stylized}
\vspace{-0mm}
\end{table}

\begin{table}[t!]
\begin{centering}
\includegraphics[width=0.95\columnwidth]{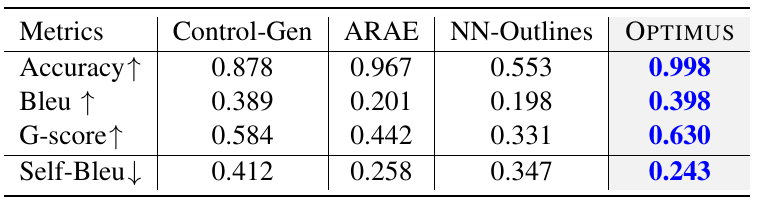}
\end{centering}
\vspace{-2mm}
\caption{Label-conditional text generation on  $\mathtt{Yelp}$.}
\label{table:label_condtional}
\vspace{-0mm}
\end{table}



\begin{table*}[t!]
    \centering
  \vspace{-0mm}
\includegraphics[width=2.06\columnwidth]{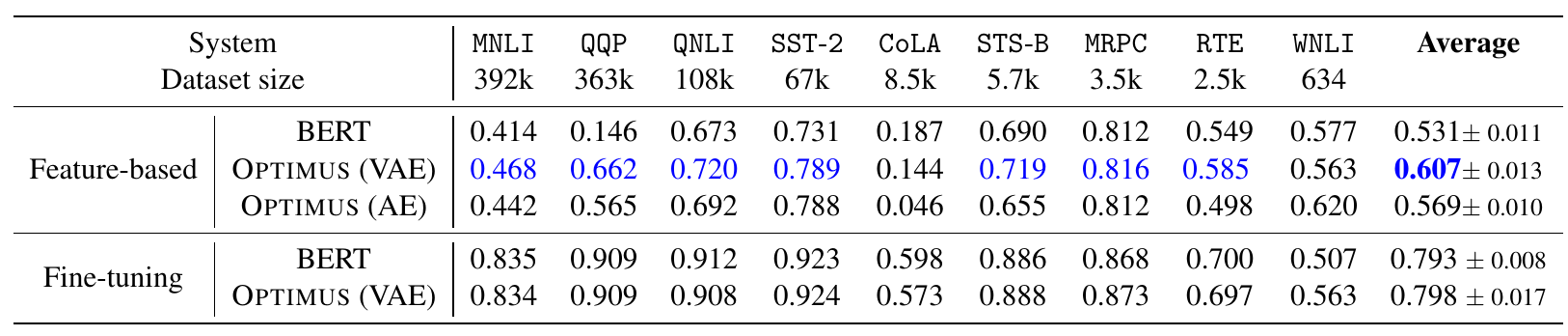}
\vspace{-0mm}
  \caption{Comparison of BERT and \short{} (with the AE and VAE objectives). Comparison is on the validation set of GLUE. F1 scores are reported for $\mathtt{QQP}$ and $\mathtt{MRPC}$, Spearman correlations are reported for  $\mathtt{STS\text{-}B}$, and accuracy scores are reported for the other tasks.}
  \label{tab:compare_glue}
\vspace{-0mm}
\end{table*}

\paragraph{Stylized response generation} Following StyleFusion~\cite{gao2019structuring}, we consider generating responses for $\mathtt{Dailydialog}$ in the style of Holmes. The comparison is shown in Table~\ref{table:dialogue_stylized}. In addition to Bleu, we use neural and N-gram classifier scores to evaluate the accuracy of the generated responses that belong to the desired style. \short{} achieves better performance on all metrics. 

\paragraph{Label-conditional text generation} 
The short $\mathtt{Yelp}$ dataset collected in~\cite{shen2017style} is used. It contains 444K training sentences, and we use separated datasets of 10K sentences for validation/testing, respectively. The goal is to generate text reviews given the positive/negative sentiment. %
We fine-tune \short{} using the VAE objective on the dataset, then freeze backbone weights. A conditional GAN~\cite{mirza2014conditional} is trained on the fixed latent space. The generation process is to first produce a latent vector $\zv_{y}$ based on a given label $y$ using conditional GAN, then generate sentences conditioned on $\zv_y$ using the decoder. 
The baselines are described in Appendix. G-score computes the geometric mean of Accuracy and Bleu, measuring the comprehensive quality of both content and style.
Self-Bleu measures the diversity of the generated sentences. The results are shown in Table~\ref{table:label_condtional},~\short{} achieves the best performance on all metrics. 
This verifies the importance of learning a smooth and meaningful latent space.
The conditional generated sentences are shown in Appendix.

\subsection{Low-resource Language Understanding}
Due to the regularization term $\Lcal_{R}$, \short{} can organize sentences in the way specified by the prior distribution. For basic VAEs, a smooth feature space is learned, which is specifically beneficial for better generalization when the number of task-specific labeled data is low. 
To have a fair comparison, we follow the BERT paper, where the hidden feature of $\texttt{[CLS]}$ is used as the
sentence-level representation. In this way, the linear classifiers for both models have the same number of trainable
parameters. Though the latent vector $\zv$ is typically used as sentence-level representation in VAE literature, we argue that the KL regularization applied on $\zv$ has a large impact on the preceding layer feature $\hv_{ \texttt{[CLS]}}$.
Specifically, $\hv_{ \texttt{[CLS]}}$ is fed into an linear classifier $\Wmat_{\text{C}} \in \R^{K \times H}$, where $K$ is the number of classes, with objective $-\log (\text{softmax} (\hv_{ \texttt{[CLS]}} \Wmat_{\text{C}}^{\top} ) )$. Two schemes are used: 
$(\RN{1})$  {\it Fine-tuning}, where both the pre-trained model and the classifier are updated; 
$(\RN{2})$  {\it Feature-based}, where pre-trained model weights are frozen to provide embeddings for the classifier update. 

\begin{figure}[t!]
	\vspace{-0mm}\centering
	\includegraphics[width=7.25cm, height=4.95cm]{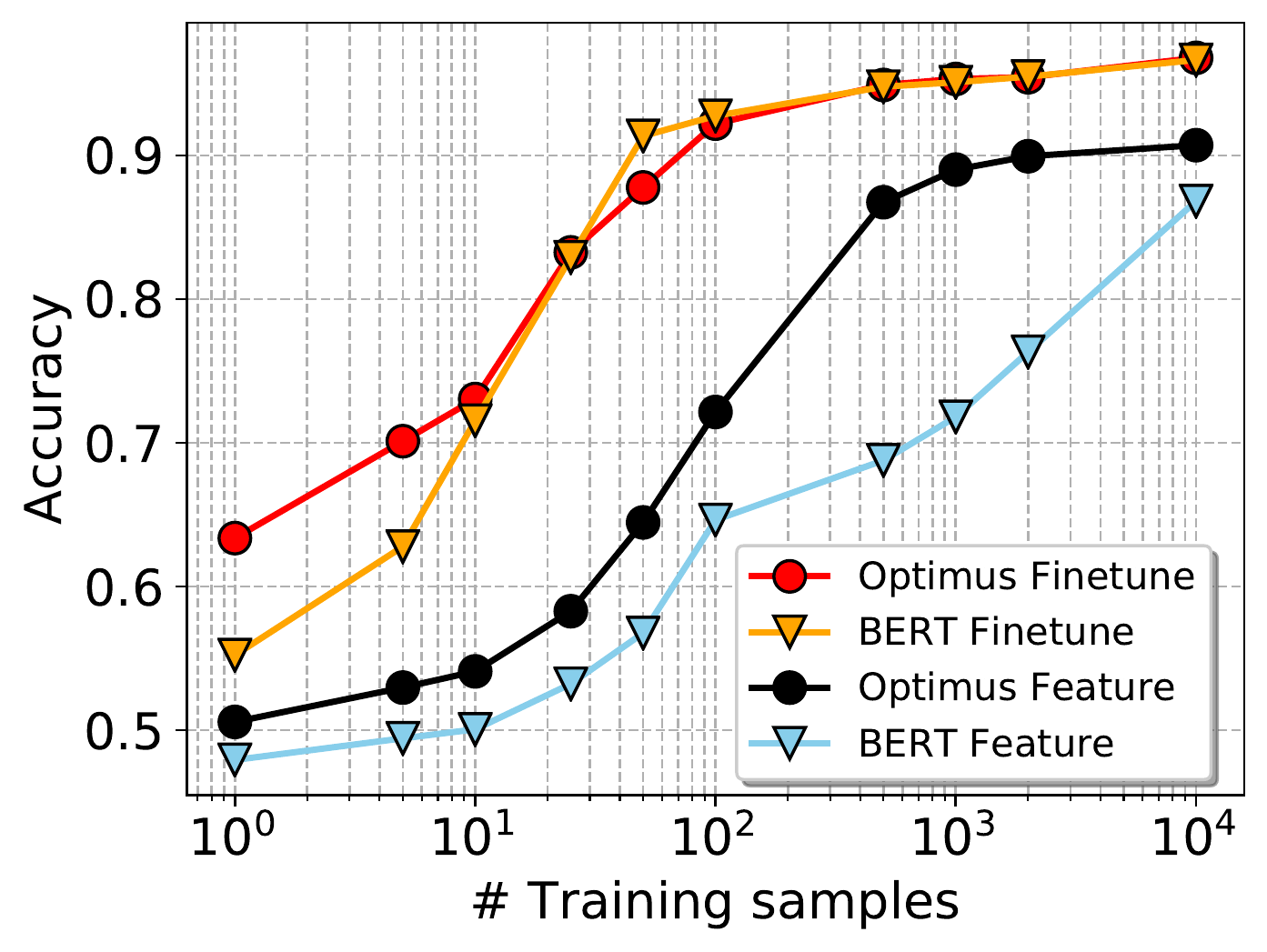}
	\vspace{-2mm}
	\caption{Testing accuracy with a varying number of labeled training samples per class on the $\mathtt{Yelp}$ dataset.}
	\label{fig:ssl_yel_acc}
	\vspace{-0mm}
\end{figure}

\paragraph{Sentiment classification on Yelp dataset.} 
A varying number of training samples are randomly chosen, ranging from 1 to 10K per class. 10 trials are used when the number of available training samples are small, each is trained in 100 training epochs. 
The results are shown in Figure~\ref{fig:ssl_yel_acc}. 
When pre-trained models are used to provide sentence embeddings, the proposed \short{}  consistently outperforms BERT. It demonstrates that the latent structure learned by \short{}  is more separated, and helps generalize better. When the entire network is fine-tuned, \short{} can adapt faster than BERT, when the available number of training samples is small. The two methods perform quite similarly when more training data is provided. This is because the pre-trained backbone network size is much larger than the classifier, where the performance is dominated by the backbone networks.

\begin{figure}[t!]
	\vspace{-0mm}\centering
	\begin{tabular}{cc}
	    \hspace{-3mm}		
		\includegraphics[height=3.8cm]{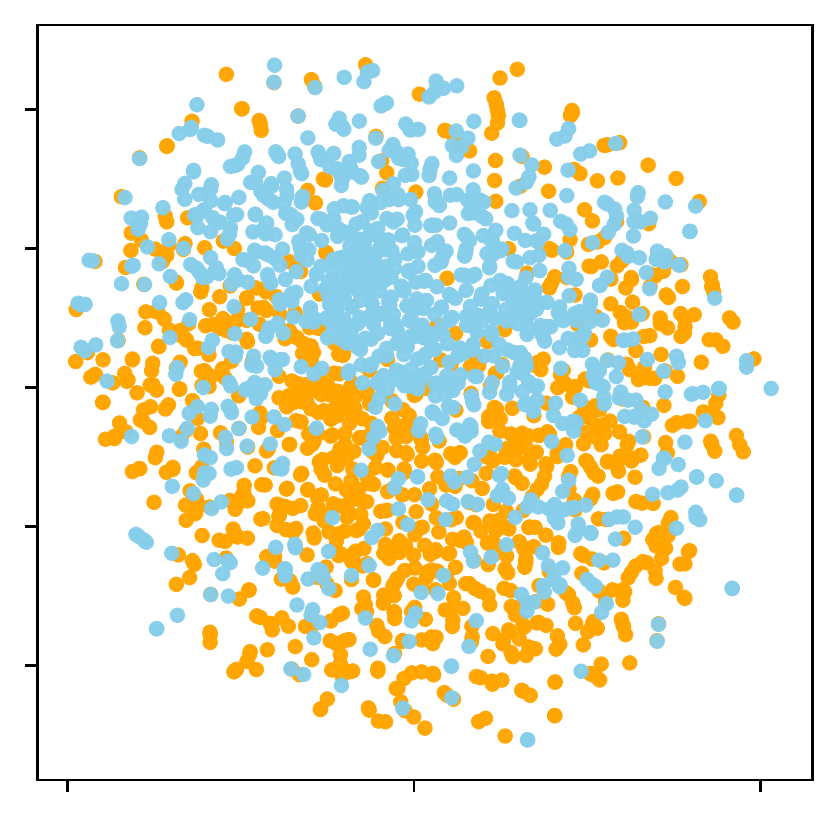} &
		 \hspace{-6mm}	
		\includegraphics[height=3.8cm]{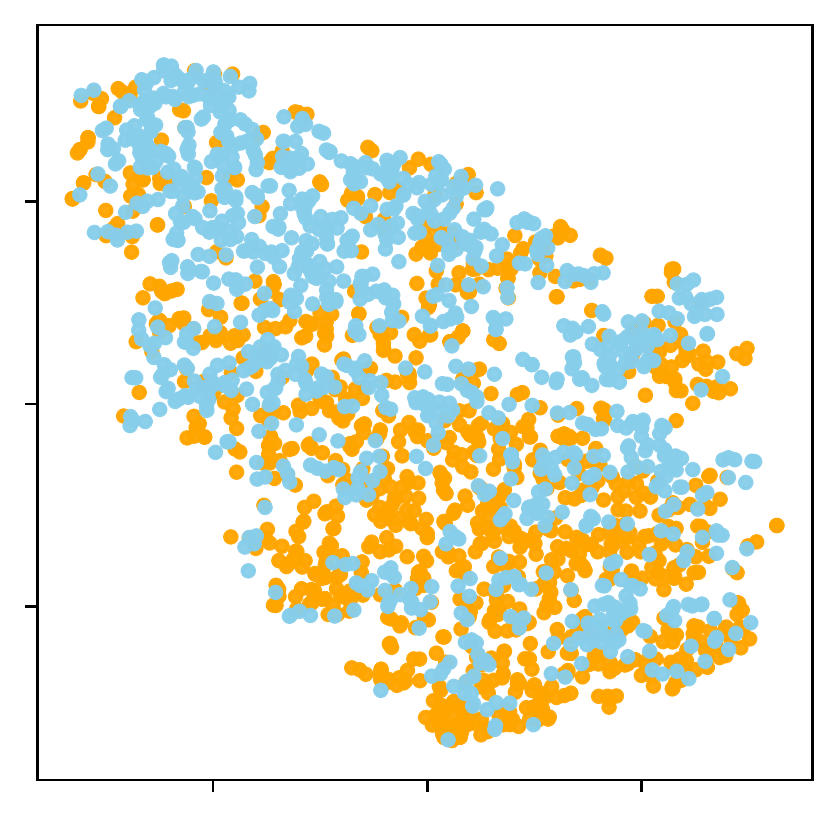} \\
		\hspace{-0mm}		
		(a) \short{}  &
		\hspace{-6mm}	
		(b) BERT
	\end{tabular}
	\vspace{-2mm}
	\caption{Comparison of tSNE visualization for the learned features. The colors indicate different labels.}
	\vspace{-2mm}
	\label{fig:tsne_latent_supp}
\end{figure}

\paragraph{Visualization of the latent space.}
We use tSNE~\cite{maaten2008visualizing}  to visualize the learned feature on a 2D map. The validation set of Yelp is used to extract the latent features.
Compared with BERT, \short{}  learns a smoother space and more structured latent patterns, which explains why \short{}  can yield better classification performance and faster adaptation.

\paragraph{GLUE.} We further consider the GLUE benchmark~\cite{wang2018glue}, which consists of nine datasets for general language understanding.
Following the finetuning schedule in~\cite{devlin2019bert}, we use learning rate $[2,3,4,5] \times 10^{-5}$ and train the model for 3 epochs. We select the best performance among different runs. We show the results on the validation set in Table~\ref{tab:compare_glue}.
With the feature-based scheme, \short{}  yields higher performance than BERT, especially on the large datasets such as MNLI, QQP and QNLI. When the full models are fine-tuned, the two methods perform quite similarly.

In summary, the scenarios that \short{} fit the low-resource settings are two-fold: (1) The required computing resource is low: the feature-based approach only updates the classifier, whose
computing requirement is much lower than full-model fine-tuning; (2) The number of required labelled data is low: when labelled data is rare, \short{} adapts better.
The results confirm that \short{} can maintain and exploit the  structures learned in pre-training, and presents a more general representation that can be adapted to new tasks more easily than BERT -- feature-based adaption is much faster and easier to perform than fine-tuning. 
\section{Discussion} 

We present \short, a large-scale pre-trained deep latent variable model for natural language. It introduces a smooth and universal latent space, by combining the advantages of VAEs, BERT and GPT-2 in one model. Experimental results on a wide range of tasks and datasets have demonstrated the strong performance of \short{}, including new state-of-the-art for language VAEs.

There are several limitations in current \short{}. First, our pre-trained language VAE is still under-trained due to limited compute resource, as the training reconstruction loss can still decrease. One may further train the models with higher latent dimension and longer time to fully release the power of pre-trained latent spaces. Second, the current model can only control sentences of moderate length. One future direction is to consider more sophisticated mechanisms to gain stronger control-ability over longer sentences while maintaining the compactness of latent representations.

While deep generative models (DGMs) such as VAEs are theoretically attractive due to its principle nature, it is now rarely used by practitioners in the modern pre-trained language modeling era where BERT/GPT dominate with strong empirical performance. That's why this paper makes a timely contribution to making DGMs practical for NLP. We hope that this paper will help renew interest in DGMs for this purpose.
Hence, we deliberately keep a simple model, believing that the first pre-trained big VAE model itself and its implications are novel: it helps the community to recognize the importance of DGMs in the pre-training era, and revisit DGMs to make it more practical.
Indeed, \short{} is uniquely positioned to learn a smooth latent space to organize sentences, which can enable guided language generation compared with GPT-2, and yield better generalization in low-resource language understanding tasks than BERT.

\section*{Acknowledgments}
The authors gratefully acknowledge Jason Yosinski, Changyou Chen, Yang Zhao and Le Fang for helpful discussion.
Additional thanks go to the entire Project Philly team inside Microsoft, who provided us the computing platform for our research. The implementation in our experiments depends on open source GitHub repositories; we acknowledge all the authors who made their code public, which tremendously accelerates our project progress.
 
\bibliography{emnlp2020} 
\bibliographystyle{acl_natbib}

\newpage
~
\newpage
\appendix

\section{Information Bottleneck and VAEs}
\label{supp_sec:ib_proof} 
\paragraph{Definition of IB}
\citet{tishby2000information} presented the Information Bottleneck (IB) method via solving the Lagrange relaxation of the optimization problem: 
\begin{align}
\hspace{-5mm}
\min \Lcal_{\text{IB}} & =  -I(\zv; \Tilde{\xv}) + \beta I(\zv; \xv)
\label{eq_ib} 
\end{align}
where $\zv$ is the representation of $\xv$, and $\beta$ is a positive parameter that controls the trade-off between the compression of input $\xv$ and preserved information about target $\Tilde{\xv}$.

In the following, we first show that the KL and reconstruction terms of VAE are the bounds of MI, respectively. Further, we put the bounds together, and show that VAE objective can optimize IB.

\paragraph{KL upper bounds MI}
Following~\cite{makhzani2016adversarial}, we refer to $q(\zv) = \int_{\xv} q(\zv | \xv) q(\xv) d \xv $ as the aggregated posterior. This marginal distribution captures the aggregated $\zv$ over the entire dataset.
The KL term~\eqref{eq_kl} in can be decomposed into two refined terms~\cite{chenisolating,hoffman2016elbo}: 
\vspace{-1mm}
\begin{align} \label{eq_kl_decomp}
 \hspace{-4mm} \Fcal_R & = \E_{q(\xv)}[ \mbox{KL} (q(\zv | \xv) || p(\zv) ) ]  \nonumber \\
& =  \underbrace{I_q(\zv, \xv)}_{\Fcal_1:~\text{Mutual~Info.}} +
\underbrace{ \mbox{KL}(q(\zv) || p(\zv)) }_{\Fcal_2:~\text{Marginal KL} }  \\
& \ge I_q(\zv, \xv)  \nonumber 
\end{align}
where 
$\Fcal_1$ is the mutual information (MI) measured by $q$. Higher MI can lead to a higher correlation between the latent variable and data variable, and encourages a reduction in the degree of KL vanishing. The marginal KL is represented by $\Fcal_2$, and it measures the fitness of the aggregated posterior to the prior distribution.

\paragraph{Reconstruction lower bounds MI}
The reconstruction term in~\eqref{eq_rec} provides a lower bound for MI measured by $q$, based on Corollary 3 in~\cite{li2017alice}:
\vspace{-2mm}
\begin{align} \label{eq_rec_bound}
 \hspace{-2mm} \Fcal_E 
 & = E_{q(\xv), \zv \sim q(\zv|\xv)} (\log p(\Tilde{\xv} | \zv ) ) ] + H_q(\Tilde{\xv}) \nonumber \\
 & \le I_q(\zv, \Tilde{\xv})
\end{align}
where $\Tilde{\xv}$ is the reconstruction target in our auto-encoder setting, and $H( \Tilde{\xv} )$ is a constant. 

\paragraph{VAE recovers BI}
When scheduled with $\beta$, the training objective over the dataset can be written as:
\begin{align} \label{eq_ml_analysis}
\Fcal_{\beta} & = -\Fcal_E + \beta \Fcal_R  \\
& \ge - I_q(\zv, \Tilde{\xv} ) +  \beta I_q(\zv, \xv)
\end{align}

This recovers IB principle in \eqref{eq_ib}.  
When $\beta=0$, we have the AE variant of our \short{}, the model fully focuses on maximizing the MI to recover sentence from the latent space. As $\beta$ increases, the model gradually transits towards fitting the aggregated latent codes to the given prior, leading the VAE variant of our \short{}.

\section{Pre-training Details}
\label{sec:pre_training}

\subsection{Latent Vector Injection Schemes} 
\label{supp_sec:inject}
\begin{figure*}[t!]
	\vspace{-0mm}\centering
	\begin{tabular}{c c}
		&
		\hspace{-100mm}
		\includegraphics[height=0.7cm]{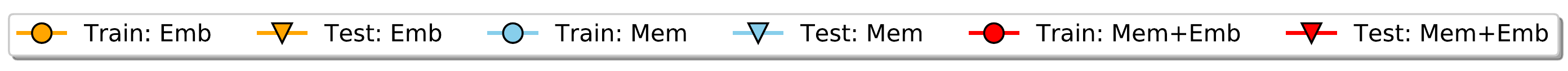}
		 \hspace{-20mm}
		 \\
		\hspace{-5mm}
		\includegraphics[height=3.6cm]{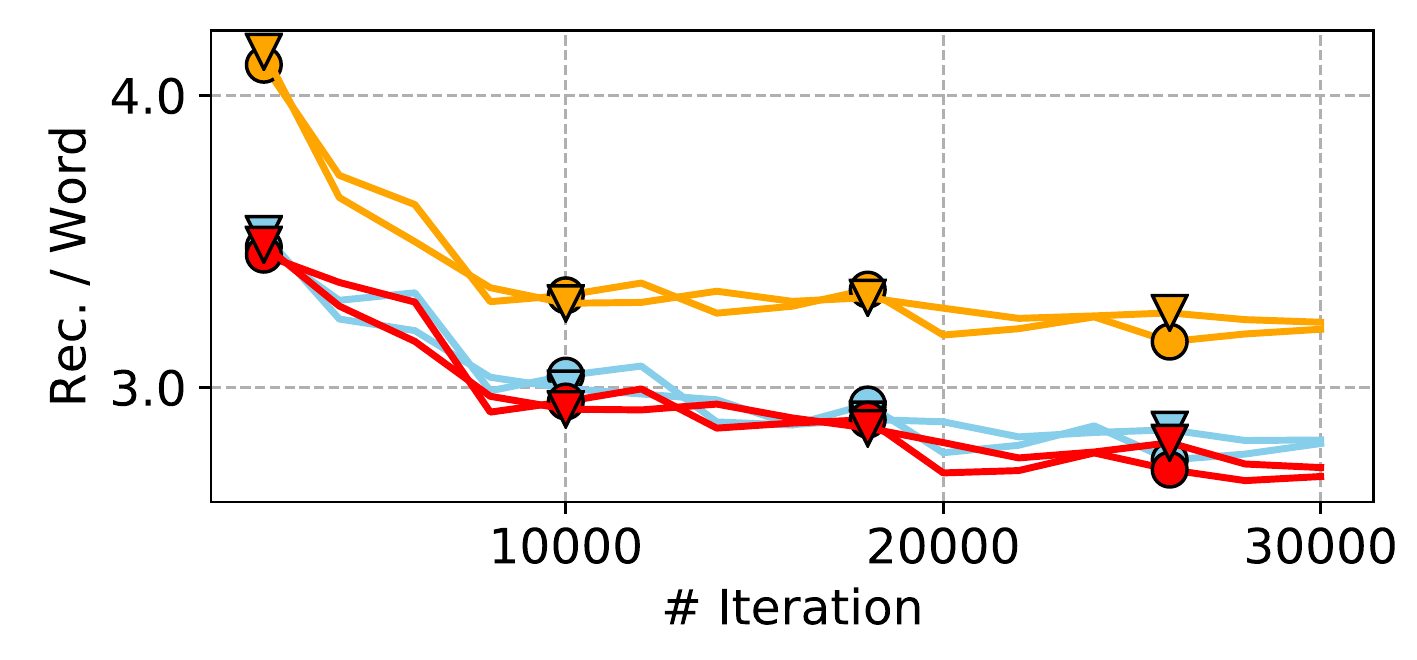}
		&
		\includegraphics[height=3.6cm]{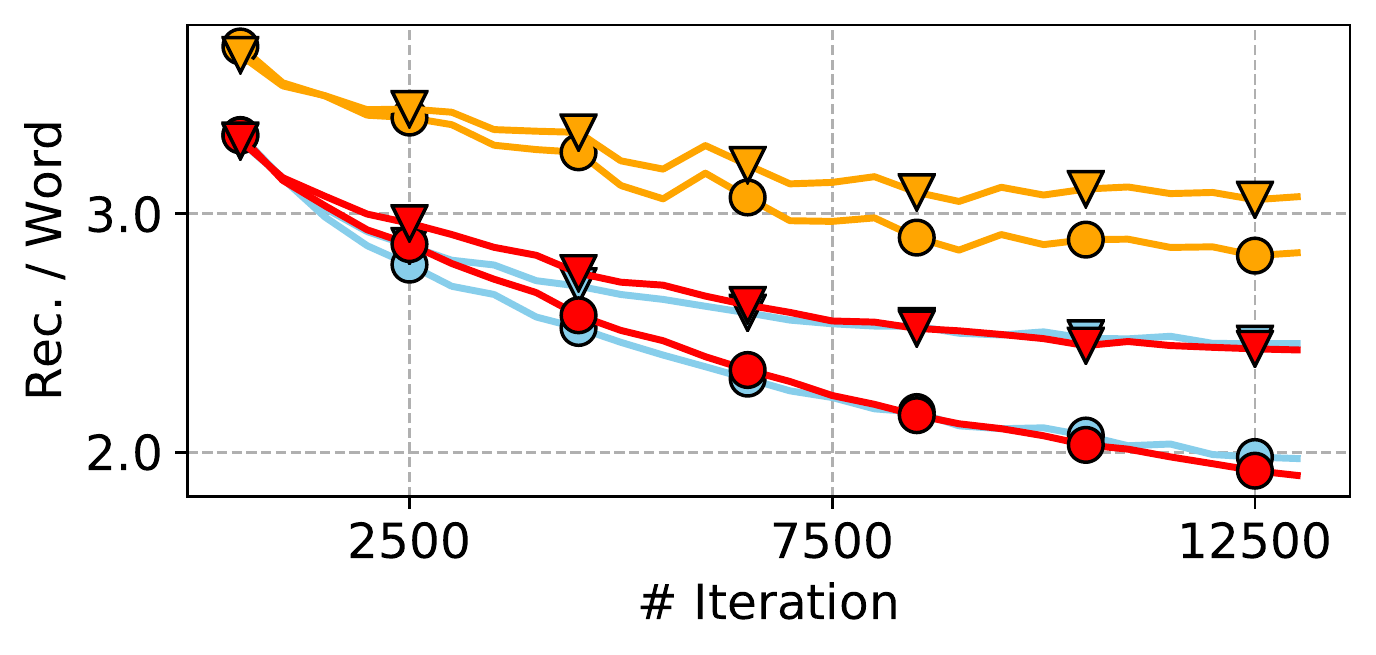} \\
		(a) Yelp \vspace{2mm} 
		&
		(b) PTB \hspace{-0mm}  \\
	\end{tabular}
	\vspace{-2mm}
	\caption{Illustration of three different schemes to inject latent vector into GPT-2 for guided language generation: (a) Yelp and (b) PTB.
	The learning curves for reconstruction error per word is considered. \texttt{Emb} indicates latent vector is used as additional embedding to add into other embeddings, and \texttt{Mem} indicates latent vector is used as additional memory token for GPT2 to attend. \texttt{Mem+Emb} indicates the integration of two schemes.
	 }
	\vspace{-0mm}
	\label{fig:schemes_inject_latent}
\end{figure*}

We compare three different schemes to inject latent vector into GPT2 in Figure~\ref{fig:schemes_inject_latent}: 

\begin{itemize}
    \item {\bf Mem}.~Latent vector $\zv$ is used as additional {\em memory} token for GPT2 to attend.
    \item {\bf Emb}.~Latent vector $\zv$ is used as additional {\em embedding} to add into other embeddings. 
    \item {\bf Mem+Emb}.~The integration of the above two schemes.
\end{itemize}
On both Yelp and PTB datasets, 5 training epochs are considered. Yelp generally has longer sentences than PTB. The encoder is initialized with BERT, and decoder is initialized with GPT-2. Lower reconstruction error per word indicates a more effective approach to pass the information flow from encoder to decoder. We see that it is significantly more efficient to use $\zv$ as a memory vector for GPT-2 to attend, than as the additional embedding. The combined scheme yields slightly better performance in the late stage of training. In the paper, we use the combined scheme in default.

\subsection{Wikipedia Dataset} 
\label{supp_sec:wiki}
\begin{figure*}[t!]
	\vspace{-0mm}\centering
	\begin{tabular}{c c}
		\hspace{-0mm}
		\includegraphics[height=3.8cm]{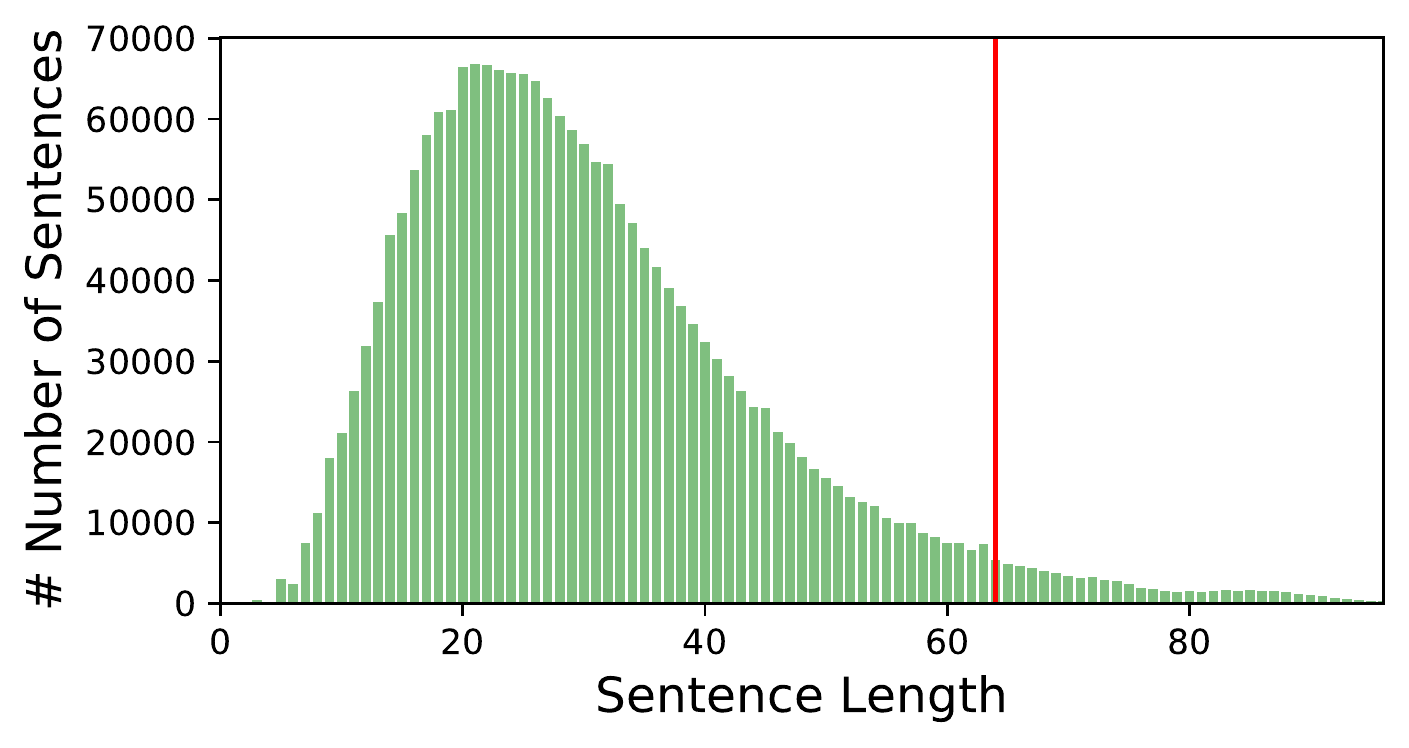}  & 
		\includegraphics[height=3.8cm]{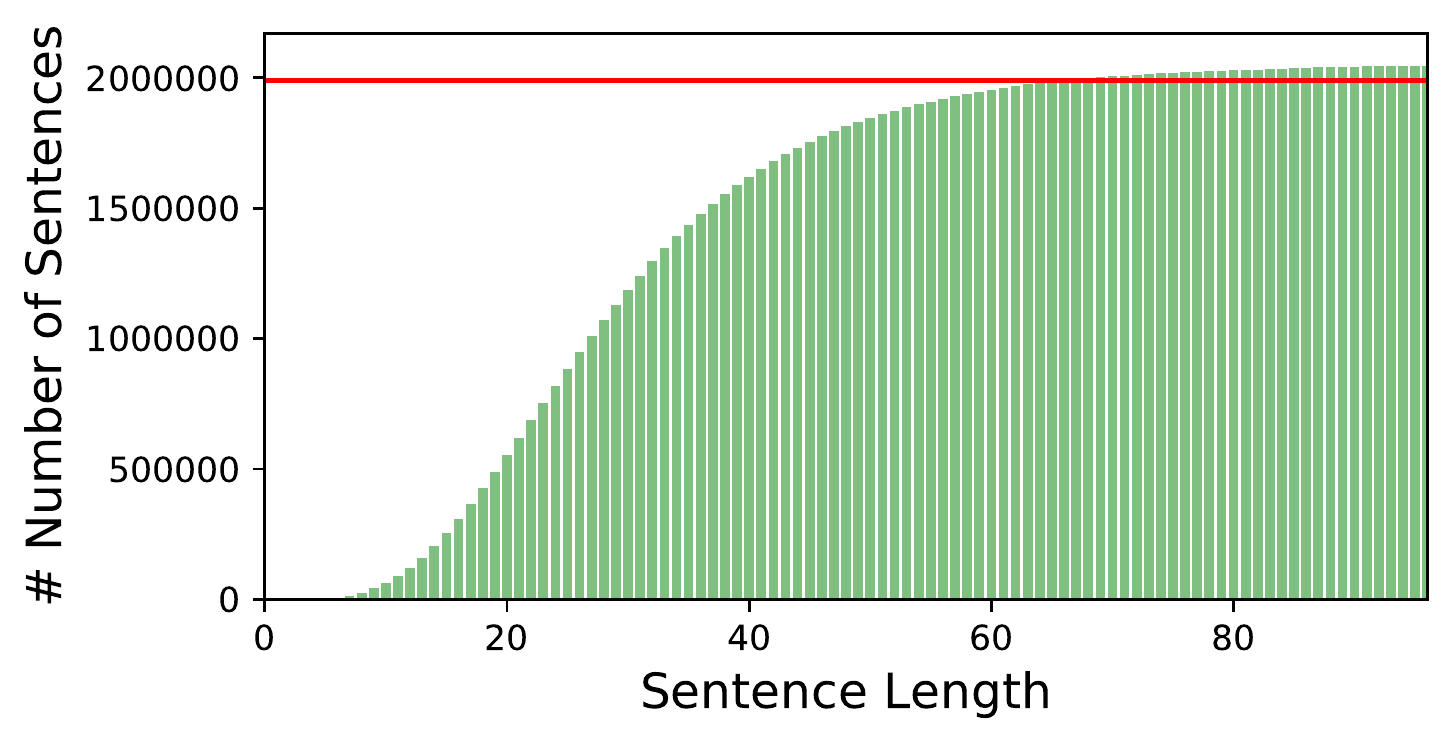} \\
		(a) Frequency distribution \vspace{2mm} & 
		(b) Cumulative frequency distribution \hspace{-0mm} \\ 
	\end{tabular}
	\vspace{-2mm}
	\caption{Illustration of sentence distribution in Wikipedia dataset: (a) Frequency distribution and (b) Cumulative Frequency distribution. We choose maximum length as 64 to construct the pre-training dataset. It leads to 1990K sentences, which is 96.45\% of entire Wikipedia dataset.
	 }
	\vspace{-0mm}
	\label{fig:wiki_stats}
\end{figure*}

We illustrate the statistics of Wikipedia dataset in Figure~\ref{fig:wiki_stats}. Since we focus on modeling natural sentences (rather than text sequences of a fixed length as in GPT-2~\cite{radford2019language}) in a latent space, we pre-process Wikipedia into a set of natural sentences, with maximum sequence length as 64. This leads to 1990K sentences, which is 96.45\% of entire Wikipedia dataset.

\section{Experiment Details}
\subsection{Language Modeling}
In addition to generating high-quality sentences as in the traditional language models that only, VAEs also aim to learn a good posterior distribution in the latent space.
The language modeling performance is evaluated with ELBO, perplexity (PPL) or importance weighted perplexity~\cite{he2019lagging}, which provides a tighter bound to $\log p(\xv)$. Higher ELBO and lower PPL indicate the model fits the observed sentences better. The pre-training takes around 50 hours for one epoch on eight V100 DGX2 GPU's.

\begin{itemize}
    \vspace{-2mm}
    \item {\bf ELBO}: The sum of KL divergence and reconstruction loss.
\item  {\bf Perplexity.} $\text{PPL} = p(x_1, \cdots, x_N)^{-1/N}$, where $N$ is the number of words. For latent variable models, we use a lower bound on the marginal log-likelihood $\log p(\xv)$, as follows from Jensen's Inequality and the
fact that the average importance weights are an unbiased estimator of $p(\xv)$:
\begin{align}
\hspace{-5mm}
\Lcal_{k} & = \E  \big[\log \frac{1}{k} \sum_{i=1}^{k} w_i  \big]  \nonumber \\
&
\le 
\log  \big[ \E  \frac{1}{k} \sum_{i=1}^{k} w_i  \big]
= \log p(\xv).
\label{eq_iw_bound} 
\end{align}
\text{where}~ $w_i = p(\xv, \zv_i)/q(\zv_i | \xv)$.
\end{itemize}

More importantly, we are interested in the learned $\zv$, which is evaluated using the following three metrics: 
\vspace{-2mm}
\begin{itemize}
    \item  {\bf AU}: The total number of active units in $\zv$, defined as $A_z = \mbox{Cov}_{\xv}(\E_{z\sim q(z|\xv)}[z])>0.01$~\cite{burda2015importance};
    \item {\bf MI}: The mutual information $I(\xv, \zv)$; 
    \item {\bf KL}: The posterior-prior KL divergence
\end{itemize}




\begin{table*}[t!]
    \centering
  \vspace{-0mm}
  \begin{tabular}{c|ccc|ccc|ccc}
    \toprule
     Metric &
        \multicolumn{3}{ c|}{LM}  & 
        \multicolumn{3}{ c|}{Representation}  & 
        \multicolumn{3}{ c }{Learning Objective}  \\ \hline
     Method
         & PPL $\downarrow$ &  &  
         & MI  $\uparrow$ & AU $\uparrow$ &  
         & -ELBO  $\downarrow$ & KL  $\uparrow$ & Rec $\downarrow$   \\
    \midrule       
    Ours($\lambda\!=\!0.05$)         
         & 23.58 &  & 
         & 3.78 & 32 & 
         & 91.31 & 4.88 &  86.43 \\ 
    Ours($\lambda\!=\!0.1$)         
         & 23.66 &  & 
         & 4.29 & 32 & 
         & 91.60 & 5.82 & 85.78  \\ 
    Ours($\lambda\!=\!0.25$)         
         & 24.24 &  & 
         & 5.98 & 32 & 
         & 93.18 & 9.42 & 83.75  \\            
    Ours($\lambda\!=\!0.5$)         
         & 26.69 &  & 
         & 7.64 & 32 & 
         & 96.82 & 15.72 & 81.09  \\      
    Ours($\lambda\!=\!1.0$)         
         & 35.53 &  & 
         & 8.18 & 32  & 
         & 77.65 & 28.50 & 77.65  \\ 
        \hline
    GPT-2
         & 24.23 &  & 
         &  &  & 
         &  &  &   \\ 
         \hline
    LSTM-LM     
         & 100.47 &  & 
         &  &  & 
         & 101.04 &  &   \\ 
    LSTM-AE
         &  &  & 
         & 8.22 & 32 & 
         &  &  & 70.36  \\ 
    M. Annealing         
         & 101.40 &  & 
         & 0.0 & 0 & 
         & 101.28 & 0.0 &  101.28  \\ 
    C. Annealing 
         & 108.81 &  & 
         & 1.27 & 5 & 
         & 102.81 & 1.37 & 101.85  \\ 
    Aggressive
         & 99.83 &  & 
         & 0.83 & 4 & 
         & 101.19 & 0.93  &  100.26  \\ 
    AE-BP ($\lambda\!=\!5$)
         & 96.86 &  & 
         &  5.31 & 32 & 
         & 102.41 & 6.54 &  95.87  \\         
  \bottomrule
\end{tabular}
\vspace{-0mm}
  \caption{\small Comparison on PTB dataset.}
  \label{tab:compare_sota_penn}
\vspace{-0mm}
\end{table*}

\begin{table*}[t!]
    \centering
  \vspace{-0mm}
  \begin{tabular}{c|ccc|ccc|ccc}
    \toprule
     Metric &
        \multicolumn{3}{ c|}{LM}  & 
        \multicolumn{3}{ c|}{Representation}  & 
        \multicolumn{3}{ c }{Learning Objective}  \\ \hline
     Method
         & PPL $\downarrow$ &  &  
         & MI  $\uparrow$ & AU $\uparrow$ &  
         & -ELBO  $\downarrow$ & KL  $\uparrow$ & Rec $\downarrow$   \\
    \midrule       
    Ours($\lambda\!=\!0.01$)         
         & 21.99 &  & 
         & 2.54 & 32 & 
         & 337.41 & 3.09 & 334.31  \\     
    Ours($\lambda\!=\!0.05$)         
         & 21.99 &  & 
         & 2.87 & 32 & 
         & 337.61 & 3.73 & 333.87  \\ 
    Ours($\lambda\!=\!0.25$)         
         & 22.20 &  & 
         & 5.31 & 32 & 
         & 340.03 & 8.70 & 331.33  \\          
    Ours($\lambda\!=\!0.5$)         
         & 22.79 &  & 
         & 7.67 & 32 & 
         & 344.10 &  15.09 & 329.01  \\ 
    Ours($\lambda\!=\!1.0$)         
         & 24.59 &  & 
         & 9.13 & 32 & 
         & 353.67 &  27.89 & 325.77 \\                   
        \hline
    GPT-2
         & 23.40 &  & 
         &  &  & 
         &  &  &   \\ 
         \hline
    LSTM-LM     
         &  &  & 
         &  &  & 
         & 358.10 &  &   \\ 
    LSTM-AE
         &  &  & 
         & 9.26 & 32 & 
         &  &  & 278.76  \\ 
    SA-VAE
         &  &  & 
         & 1.7 & 8 & 
         & 355.90 & 2.80 & 353.10 \\     
    M. Annealing     
         & 40.39  &  &  
         & 0.13 & 1 & 
         & 357.76 & 0.14 & 357.62  \\ 
    C. Annealing
         &  &  & 
         &  &  & 
         &  &  &   \\   
    Aggressive
         &  &  & 
         & 2.4 & 7 & 
         & 328.40 & 3.4 & 322.70  \\ 
    AE-BP ($\lambda\!=\!5$)
         &  &  & 
         &  &  & 
         &  &  &   \\          
  \bottomrule
\end{tabular}
\vspace{-0mm}
  \caption{\small Comparison on Yelp dataset.  For LSTM-LM and GPT-2, we report the exact negative log
likelihood.}
  \label{tab:compare_sota_yelp}
\vspace{-0mm}
\end{table*}

\begin{table*}[t!]
    \centering
  \vspace{-0mm}
  \begin{tabular}{c|ccc|ccc|ccc}
    \toprule
     Metric &
        \multicolumn{3}{ c|}{LM}  & 
        \multicolumn{3}{ c|}{Representation}  & 
        \multicolumn{3}{ c }{Learning Objective}  \\ \hline
     Method
         & PPL $\downarrow$ &  &  
         & MI  $\uparrow$ & AU $\uparrow$ &  
         & -ELBO  $\downarrow$ & KL  $\uparrow$ & Rec $\downarrow$   \\
    \midrule   
    
    Ours($\lambda\!=\!0.05$)         
         & 22.34 &  & 
         & 5.34 & 32 & 
         & 282.70 & 6.97 & 282.84  \\     
    Ours($\lambda\!=\!0.10$)         
         & 22.56 &  & 
         & 5.80 & 32 & 
         & 289.88 & 7.77 & 282.11  \\ 
    Ours($\lambda\!=\!0.25$)         
         & 22.63 &  & 
         & 7.42 & 32 & 
         & 290.69 &  11.19 & 279.49  \\ 
    Ours($\lambda\!=\!0.50$)         
         & 23.11 &  & 
         & 8.85  & 32 & 
         & 293.34 &  17.45 & 275.89  \\ 
    Ours($\lambda\!=\!1.0$)         
         & 24.92 &  & 
         & 9.18 & 32 & 
         & 301.21 &   30.41 & 270.80 \\       
        \hline
    GPT-2
         & 22.00 &  & 
         &  &  & 
         &  &  &   \\ 
         \hline
    LSTM-LM     
         & 60.75 &  & 
         &  &  & 
         & 328.00  &  &   \\ 
    LSTM-AE
         &  &  & 
         & 9.26 & 32 & 
         &  &  & 278.76  \\ 
    SA-VAE
         & 60.40 &  & 
         & 2.70 & 10 & 
         &  327.20 &  5.20 & 325.00 \\     
    M. Annealing         
         & 61.21  &  &  
         & 0.0 & 0 & 
         & 328.80 & 0.0 & 328.80  \\ 
    C. Annealing
         & 64.26 &  & 
         & 0.0 & 1 & 
         & 332.68 & 0.03 & 332.65  \\
    Aggressive 
         & 59.77 &  & 
         & 2.9 & 15 & 
         & 328.40 & 5.70 & 322.70  \\ 
    AE-BP ($\lambda\!=\!5$)
         & 59.28 &  & 
         & 8.08 & 32 & 
         & 329.31 & 10.76 &  318.55  \\          
  \bottomrule
\end{tabular}
\vspace{-0mm}
  \caption{\small Comparison on Yahoo dataset.}
  \label{tab:compare_sota_yahoo}
\vspace{-0mm}
\end{table*}

\begin{table*}[t!]
    \centering
  \vspace{-0mm}
  \begin{tabular}{c|ccc|ccc|ccc}
    \toprule
     Metric &
        \multicolumn{3}{c|}{LM}  & 
        \multicolumn{3}{ c|}{Representation}  & 
        \multicolumn{3}{ c }{Learning Objective}  \\ \hline
     Method
         & PPL $\downarrow$ &  &  
         & MI  $\uparrow$ & AU $\uparrow$ &  
         & -ELBO  $\downarrow$ & KL  $\uparrow$ & Rec $\downarrow$   \\
    \midrule     
    %
    Ours($\lambda\!=\!0.05$)         
         & 13.47 &  & 
         & 3.49 & 32 & 
         & 33.08 & 3.92 & 29.17  \\ 
    Ours($\lambda\!=\!0.10$)         
         & 13.48 &  & 
         & 4.65 & 32 & 
         & 33.45 & 5.44 & 28.01 \\
    Ours($\lambda\!=\!0.25$)         
         & 14.08 &  & 
         & 7.22 & 32 & 
         & 35.04 & 9.79 & 25.25 \\    
    Ours($\lambda\!=\!0.50$)         
         & 16.67 &  & 
         & 8.89 & 32 & 
         & 38.50 & 16.35 & 22.14 \\    
    Ours($\lambda\!=\!1.00$)         
         & 29.63 &  & 
         & 9.20 & 32 & 
         & 47.35 & 28.96 & 18.39  \\  
         \hline         
    GPT-2~\cite{radford2019language}
         & 20.24 &  & 
         &  &  & 
         &  &  &   \\ 
         \hline
    LSTM-LM     
         &  21.44 &  & 
         &  &  & 
         &  &  &   \\ 
    LSTM-AE
         &  &  & 
         & 9.18 & 32 & 
         &  &  &  \\ 
    M. Annealing~\cite{bowman2015generating}         
         & 21.50  &  &  
         &  1.42 & 2 & 
         & 33.07 &  1.42 & 31.66  \\ 
    C. Annealing~\cite{fu2019cyclical} 
         & 21.62 &  & 
         &  2.33 & 4 & 
         & 33.25 &  2.36 & 30.89  \\   
    Aggressive~\cite{he2019lagging} 
         & 21.16  &  & 
         & 1.38 & 5 & 
         & 32.95 & 1.42 & 31.53  \\ 
    AE-BP ($\lambda\!=\!5$)~\cite{li2019surprisingly} 
         & 21.64  &  & 
         & 7.71 & 32 & 
         & 34.47 &  9.53 &  24.94  \\          
  \bottomrule
\end{tabular}
  \caption{\small Comparison on SNLI dataset.  For LSTM-LM and GPT-2, we report the exact negative log
likelihood.}
  \label{tab:compare_sota_snili}
\vspace{10mm}
\end{table*}

The full experimental results on shown in Table~\ref{tab:compare_sota_penn}, \ref{tab:compare_sota_yelp}, \ref{tab:compare_sota_yahoo} and \ref{tab:compare_sota_snili}.

\subsection{Dialog response generation} 
\paragraph{Dialog response generation: SpaceFusion}
We interpolate samples $\zv_{\tau}$ between the context and response as $\zv_{\tau} = \tau \zv_{\text{S2S}} +  (1 - \tau) \zv_{\text{AE}}$, where $\tau \sim \text{Uniform} (0,1)$. 
We fix the first 11 layers of encoder, and fine-tune from last layer to $\zv$: $\{{\phiv_{\text{AE}}},  \phiv_{\text{E}}\}$.  An additional network path $\{{\phiv_{\text{S2S}}},  \phiv_{\text{E}}^{\prime} \}$ is introduced from the 11th layer of encoder to $\zv$ to represent context. The fine-tuning objective is:
$$ \min_{ \{\phiv_{\text{S2S}}, \phiv_{\text{AE}}, \phiv_{\text{E}}, \phiv_{\text{E}}^{\prime}, \thetav \} } \Lcal_{{\text{dialog}}} = 
\Lcal_{{\xv}} + \Lcal_{{\text{fusion}}}
$$ 
where $ \Lcal_{{\text{fusion}}}$ is the same with fusion term in~\cite{gao2019jointly}, and  $ \Lcal_{{\xv}} = 
- [\log p (\xv | \zv_{\text{S2S}}  ) 
+ \log p (\xv | \zv_{\text{AE}} )
+ \log p (\xv | \zv_{\tau} ) ]
$.

We benchmark representative baselines and state-of-the-art approaches, including: 
$(\RN{1})$ Seq2Seq: a generalized sequence-to-sequence model with hierarchical RNN encoder~\cite{serban2016building};
$(\RN{2})$ SeqGAN: a GAN based model for sequence generation~\cite{li2017adversarial}; 
$(\RN{3})$ CVAE baseline \cite{zhao2017learning}; 
$(\RN{4})$ Dialogue WAE, a conditional Wasserstein auto-encoder for response generation \cite{gu2018dialogwae};
$(\RN{5})$: A hierarchical VAE model~\cite{serban2017hierarchical}. 
$(\RN{6})$ VHCR: a hierarchical VAE model with conversation modeling~\cite{park2018hierarchical}.
$(\RN{7})$ iVAE$_{\text{MI}}$: An implicit VAE model augmented with mutual information regularizer~\cite{fang2019implicit}.
The full comparison in shown in Table~\ref{table:dialogue_appendix}.

\paragraph{Stylized response generation: StyleFusion} In this task, the additional sentences $\bv$ are used to bias the generated response towards the reference style. The biased response representation is $\zv_{\tau}^{\prime} = \tau \zv_{\text{Style}} +  (1 - \tau) \zv_{\text{AE}}$, where $\tau \sim \text{Uniform} (0,1)$ and $\zv_{\text{Style}}$ is the latent representation of $\bv$. The corresponding loss for the biased target is 
$\Lcal_{\xv}^{\prime} = - [\tau \log p (\xv | \zv_{\text{Style}} ) +   (1 - \tau) \log p (\xv | \zv_{\text{AE}} ) ]$, which is added into $\Lcal_{{\text{dialog}}}$ for training.

\paragraph{Evaluation}  Two type of {\it Accuracy} are reported, based on text sequence (\ie neural) and its N-gram information. The accuracy is assessed by an oracle classifier to correctly predict whether generated response belongs the style-reference dataset. 

\subsubsection{Label-Conditional Text Generation} 
The goal of this task is to generate sentences conditioned on a given label. We consider a two-stage algorithm to adapt \short{} for this task. First, we fine-tune a VAE language model on the downstream dataset, and freeze the model parameters. In another word, the latent space is fixed. Second, we build a conditional GAN for the latent space. Let's denote the latent vectors for ground-trurh sentences as $\zv_{\text{true}}$. We build a generator $G$ to produce $\zv_{\text{fake}} = G(\epsilon, y)$, where $\epsilon$ is the random noise, and $y$ is the label. A discriminator $D$ is trained simultaneously to distinguish $\zv_{\text{true}}$ and $\zv_{\text{fake}}$. The learning objectives for conditional GAN is: 

\begin{align}
    & \min_{G} \max_{D}\mathcal{L_{\text{cGAN}} }   \nonumber \\ 
    &= 
    \mathbb{E}_{\xv, y \sim q(\xv, y)} \big[ 
    \mathbb{E}_{\zv \sim q(\zv | \xv)} [\log p_{D} (d=1|E(\xv))] \nonumber \\ 
    &+ \mathbb{E}_{\epsilon \sim p_0(\epsilon)} [\log p_{D} (d=0|G(\epsilon, y))] \big] 
    \label{eq:L_cgan}
\end{align}

To make the model work effectively, it is key to learn a smooth and meaningful latent space of target sentences.  The text generation procedure conditioned on label $y$ is:  

\begin{align}
   \xv \sim p_{\thetav}(\xv | \zv ), ~~~\text{with}~~~  \zv = G(\epsilon, y) 
    \label{eq:L_cgan_generation}
\end{align}

This mimics the process to produce the outlines of the sentences using conditional GAN, and fill in details using the decoder. We show some generated sentences in Table~\ref{table:label_conditional_example1}.

We compare with three baselines: (1) {\it Ctrl-Gen}~\cite{hu2017toward}; We use their released code to reproduce the results. 
(2) {\it ARAE} ~\cite{zhao2018adversarially} proposes to learn an auto-encoder first, and then train a GAN to produce the latent vectors.  
(3) {\it NN-Outlines} ~\cite{subramanian2018towards} proposes the use of a general purpose encoder for text generation, and we implement it using BERT. Note that our two-stage fine-tuning scheme borrows the ideas from ARAE and NN-Outlines. The key difference is that we employ our pre-trained \short{} model, and work on a better latent space. 

\paragraph{Evaluation} 
We consider three metrics: (1)  {\it Bleu} for sentence quality, (2)  {\it Accuracy} for conditional generation capability. The accuracy is assessed by an oracle classifier to correctly predict the attributes that generated sentences are conditioned on. (3) {\it G-score} is reported as the geometric mean of Accuracy and Bleu. This is the most important metric, as it evaluates the overall performance.  
For label-conditional text generation, Bleu of each generated sentence is computed by comparing with all sentences in the test set, as there are no source sentences. We further report Self-Bleu~\cite{zhu2018texygen} to evaluate the diversity of generated sentences.

\subsection{Latent space interpolation \& arithmetic operation}

\paragraph{Arithmetic operation}
The universal latent space learned by \short{} supports arithmetic operations. Given source sentence $\xv_A$ and target $\xv_B$, the goal is to re-write the input sentence $\xv_C$ as output $\xv_D$ in analogy to the transition from $\xv_A$ to $\xv_B$. We first encode $\xv_{A,B,C}$ into the latent vectors $\zv_{A,B,C}$, respectively, then apply the arithmetic operator $\zv_D= \zv_B - \zv_A + \zv_C$, and generate $\xv_D$ conditioned on $\zv_D$. One example is shown in Table~\ref{table:transfer_example_main}. Interestingly, we observe consistent style transfer from  $\xv_C$ to $\xv_D$ , to analogize the relation from $\xv_A$ to $\xv_B$. For example, the subject is revised from singular to plural forms, the topic changes from daily-life to sport. In another word, \short{} supports sentence arithmetic operator $\xv_D \approx \xv_B - \xv_A + \xv_C$ at the semantic level. More latent vector arithmetic operation examples are shown in Table~\ref{table:transfer_example1},~\ref{table:transfer_example2},~\ref{table:transfer_example3}.  

\paragraph{Latent space interpolation} 
One favorable property of VAEs is to provide a smooth space that captures sentence semantics. We demonstrate linear interpolating between latent vectors.  We take two sentences $\xv_{1}$ and $\xv_{2}$, and use their posterior mean as the latent features $\zv_{1}$ and $\zv_{2}$, respectively. We interpolate a path $\zv_{\tau}=\zv_{1}\cdot(1-\tau)+\zv_{2}\cdot \tau$ with $\tau$ increased from 0 to 1 by a step size of 0.1. Table \ref{table:interpolation_example_main} shows generated sentences using greedy decoding conditioned on $\zv_{\tau}$. The interpolated sentences exhibit smooth semantic evolution. More interpolation examples are shown in Appendix. Note that we have observed smooth \& meaningful interpolation results for almost arbitrary input sentences pairs. This demonstrates the promise that \short{} learns a universal latent space. 
More latent space interpolation examples are shown in Table~\ref{table:interpolation_example1},~\ref{table:interpolation_example2},~\ref{table:interpolation_example3}. 

\paragraph{Limitations.} 
While \short{} shows the potentials of latent-vector-based controllable language generations, it has several limitations:
(1) The compactness of latent vectors restricts the amount of encoded information, thus the model has difficulties in representing with long or complex sentences. This can be improved with more sophisticated design of latent space.  
(2) The model generates repeated interpolated sentences when intrinsic language variations are limited.
(3) When doing interpolation, though the model knows the basic trend of numbers, it does not fully understand how to count numbers; For example, it jumps from one to five, then to twenty, instead of outputting the smoothly changing numbers such as one, five, ten, fifth, twenty. 

For more user interaction with \short{}, we have released a demo website to allows users to input sentences, and the system will provide controllable generated sentences with arithmetic or interpolating operations.

\subsection{Ablation study on VAE \& AE objectives} We compare the interpolation examples in Table~\ref{table:interpolation_comparison_example1}, and generally observe that VAE can produce smoother sentences interpolation results than AE. 
We compare the two pre-training objectives on the GLUE benchmark using the feature-based approach. The results are shown in Table~\ref{tab:compare_glue}. We see that both objectives outperform than BERT on large datasets, and VAE objective performs better than AE objective. This verifies the effectiveness of smooth regularization on the latent space for the classification performance.

\begin{table*}[t!]
\begin{centering}
\includegraphics[width=1.95\columnwidth]{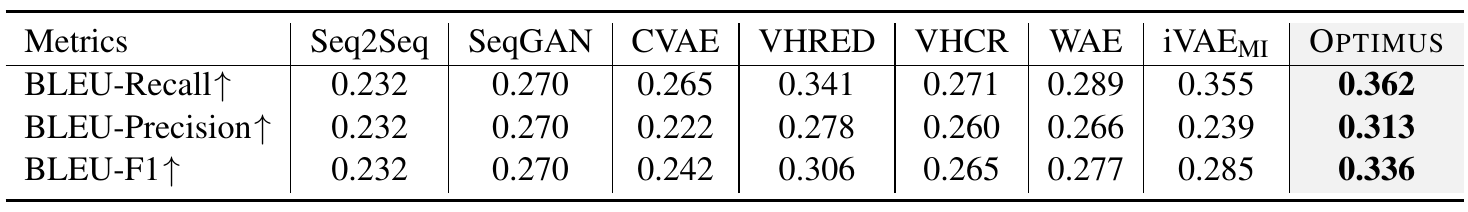}
\end{centering}
\caption{Dialog response generation on  $\mathtt{DailyDialog}$ dataset. All numbers are  from~\cite{gu2018dialogwae} except that iVAE$_{\text{MI}}$ is
from~\cite{fang2019implicit}.}
\label{table:dialogue_appendix}
\vspace{-0.0cm}
\end{table*}

\begin{table*}[t] 
    \begin{tcolorbox}[fontupper=\footnotesize, fontlower=\footnotesize]
      \hspace{-7mm}
    \begin{tabular}{l l}
$0.0$ & a young woman with a black hairbrush brushes her teeth while a man in a white shirt watches. 
\tabularnewline
$0.1$ & a blond woman with a black hairbrush brushes her teeth while a blond woman with a white hairbrush brushes her teeth.  
\tabularnewline
$0.2$ & a blond woman with a black hairbrush brushes her teeth while a man in a blue shirt watches. 
\tabularnewline
$0.3$ & a blond woman with a black hairbrush brushes her teeth while a man in a blue shirt watches.
\tabularnewline
$0.4$ & a young woman in a blue shirt and blue jeans is lifting a large plastic bottle from a bottle.
\tabularnewline
$0.5$ & a man in a blue shirt and blue jeans is brushing his teeth while a woman in a white shirt and blue pants looks on.
\tabularnewline
$0.6$ & a man in a blue shirt is holding a small plastic bag while another man in a white shirt holds a large plastic bag.
\tabularnewline
$0.7$ & a man in a blue shirt is holding a small plastic bag while another man in a white shirt holds a large plastic bag.
\tabularnewline
$0.8$ & a man in a blue shirt is holding a bag of frozen peas while another man in a white shirt looks on.
\tabularnewline
$0.9$ & a man in a blue shirt is holding a bag of food in a small bowl.
\tabularnewline
$1.0$ & a man in a blue shirt is holding a bag of food in a small area of grass.
	\end{tabular}
	\end{tcolorbox}
	\caption{Interpolating latent representation from plural sentence to singular sentence. Each row show $\tau$ and the sentence generated from the latent vector $\zv_{\tau}$. }
	\label{table:interpolation_example1}
	\vspace{2mm}
\end{table*}

\begin{table*}[t] 
    \begin{tcolorbox}[fontupper=\footnotesize, fontlower=\footnotesize]
      \hspace{-7mm}
    \begin{tabular}{l l}
$0.0$ & people are walking near a road.
\tabularnewline
$0.1$ & people are walking near a bench. 
\tabularnewline
$0.2$ & people are sitting on a bench near a road.
\tabularnewline
$0.3$ & people are sitting on a bench near a road.
\tabularnewline
$0.4$ & some people are sitting on a bench outside.
\tabularnewline
$0.5$ & there are two people sitting on a bench.
\tabularnewline
$0.6$ & there are two men sitting on a bench waiting for a train.
\tabularnewline
$0.7$ & there are two men sitting on a bench and looking at the sky.
\tabularnewline
$0.8$ & there is a man sitting on the side of a boat.
\tabularnewline
$0.9$ & there is a man sitting on the side of a boat and a woman sitting on the other side.
\tabularnewline
$1.0$ & there is a man sitting on the side of a boat and the woman is sitting on the side of a boat.
	\end{tabular}
	\end{tcolorbox}
	\caption{Interpolating latent representation from short sentence to long sentence. Each row show $\tau$ and the sentence generated from the latent vector $\zv_{\tau}$. }
	\label{table:interpolation_example2}
	\vspace{2mm}
\end{table*}

\begin{table*}[t] 
    \begin{tcolorbox}[fontupper=\footnotesize, fontlower=\footnotesize]
      \hspace{-4mm}
    \begin{tabular}{@{}p{0.8cm}@{}  @{}p{14cm}@{} }
$0.0$ & i have been here a few times and i have never had a bad experience. i ordered the chicken and waffles. the chicken was cooked perfectly and the waffles were delicious. the waffles were also very good. i would definitely come back here again. 
\tabularnewline
$0.1$ & i have been going to this place for years. i had the chicken fried rice and it was delicious. the service was great and the food was fresh. i will definitely be back. i will definitely be back.
\tabularnewline
$0.2$ & i have been going to this place for years. i was surprised to find out that they have a new location. the food is great and the service is great. i ordered the  \texttt{[UNK]} chicken and it was delicious. i also ordered the \texttt{[UNK]} chicken and it was delicious. i will definitely be back.
\tabularnewline
$0.3$ & i've been here a few times and it's always been great. the food is always fresh and the service is always fast. i'm not sure if they have a  \texttt{[UNK]} or not but i'm sure they have a  \texttt{[UNK]}. i'm sure they will be back soon.
\tabularnewline
$0.4$ & i'm not sure what to say about this place. they have a great selection of food and drinks. i had the  \texttt{[UNK]} and it was delicious. the staff was friendly and helpful. i will definitely be back.
\tabularnewline
$0.5$ & i'm not sure what to say about this place. they have a great selection of food and the staff is very friendly. i'm not sure if they have a  \texttt{[UNK]} or not. i'm sure they will be back soon.
\tabularnewline
$0.6$ & wow! this place is awesome! they have a great selection of food and the staff is very friendly. i will definitely be back.
\tabularnewline
$0.7$ & wow! this place is awesome! they have a great selection of food and the staff is very friendly. i will definitely be back.
\tabularnewline
$0.8$ & great place! they have a great selection of food. they also have a great customer service. i will definitely be back!
\tabularnewline
$0.9$ & great place! they have a great selection of products. they are very friendly and helpful. i will definitely be back!
\tabularnewline
$1.0$ & great place! they have a great customer service. they are very friendly and helpful. they are also very helpful with the  \texttt{[UNK]}. i will definitely be back!
	\end{tabular}
	\end{tcolorbox}
	\caption{Interpolating latent representation within the same sentiment. Each row show $\tau$ and the sentence generated from the latent vector $\zv_{\tau}$. }
	\label{table:interpolation_example3}
	\vspace{2mm}
\end{table*}


\begin{table*}[t!]\centering
\begin{minipage}{16cm}\vspace{0mm}    \centering
\begin{tcolorbox} 
\vspace{-2mm}
\begin{adjustbox}{scale=0.8,tabular=p{9cm} p{10cm},center}
{~~~~~~~~~~~~~~~~~~~~~\bf \short{}  (VAE, $\beta=1$) }   & {~~~~~\bf \short{} (AE, $\beta=0$) }  \tabularnewline
$\tau=0.0$~~~ the little girl plays with the toys. & 
 the little girl plays with the toys. \tabularnewline
$\tau=0.1$~~~ the child plays with the toy train. & 
 the little girl plays the playground toy. \tabularnewline
$\tau=0.2$~~~ the children play with a toy car. & 
 the children play the miniature train ride. \tabularnewline
$\tau=0.3$~~~ the children play in the ground. & 
 the children play in the museum's playground. \tabularnewline
$\tau=0.4$~~~ the children play in the playground &
 the children are watching a playhouse. \tabularnewline
$\tau=0.5$~~~ the children are playing in the playground. & 
 the children are watching a playhouse \tabularnewline
$\tau=0.6$~~~ the children are watching a play. & 
 the children are watching a playhouse \tabularnewline
$\tau=0.7$~~~ the children are watching a show. & 
 there are children watching a train. \tabularnewline
$\tau=0.8$~~~ there are children watching a circus. & 
 there are children watching a train. \tabularnewline
$\tau=0.9$~~~ there are children watching a train. & 
 there are children watching a train. \tabularnewline
$\tau=1.0$~~~ there are children watching a train. & 
 there are children watching a train.
\end{adjustbox}
\end{tcolorbox}
	\caption{Comparison of VAE and AE objective for latent space interpolation. VAE shows smoother interpolation results than AE.} 
	\label{table:interpolation_comparison_example1}
\end{minipage}
\vspace{-0mm}
\end{table*}

\begin{table*}[t!]\centering
\begin{minipage}{16cm}\vspace{0mm}    \centering
\begin{tcolorbox} 
\vspace{-2mm}
\begin{adjustbox}{scale=0.8,tabular=p{9cm} p{10cm},center}
{\bf Source} $\xv_A$ & {\bf Target}  $\xv_B$
\tabularnewline
two soccer players are playing soccer & the people are building a machine \tabularnewline
\midrule
\tabularnewline
{\bf Input} $\xv_C$  & {\bf Output} $\xv_D$  \tabularnewline
$\bullet$ people walking in the street & 
$\bullet$ \textcolor{blue}{the people were going to build the city} \tabularnewline
$\bullet$ the man was waiting for his wife to come home & 
$\bullet$ \textcolor{blue}{the man was going to get the job done} \tabularnewline
$\bullet$ two women preparing food for a table & 
$\bullet$ \textcolor{blue}{the people carefully prepared a piece of equipment} \tabularnewline
$\bullet$ two dogs chase each other through the water & 
$\bullet$ \textcolor{blue}{the vehicles get to work}
\tabularnewline
$\bullet$ a person sitting in a library reading & 
$\bullet$ \textcolor{blue}{a person working on the building}
\tabularnewline
$\bullet$ a tall human walking & 
$\bullet$ \textcolor{blue}{a construction project was made}
\tabularnewline
$\bullet$ a young boy and a young girl play in a grassy field & 
$\bullet$ \textcolor{blue}{a child is building a house for the future to see} \tabularnewline
$\bullet$ men playing music in the rain & 
$\bullet$ \textcolor{blue}{they were making a construction work}
\end{adjustbox}
\end{tcolorbox}
	\caption{Sentence transfer  via arithmetic operation in the latent space. The output sentences are in \textcolor{blue}{blue}. In this example, we see content transition from {\em relaxing} to {\em working}. }
	\label{table:transfer_example1}
\end{minipage}
\vspace{-0mm}
\end{table*}

\begin{table*}[t!]\centering
\begin{minipage}{16cm}\vspace{0mm}    \centering
\begin{tcolorbox} 
\vspace{-2mm}
\begin{adjustbox}{scale=0.8,tabular=p{9cm} p{10cm},center}
{\bf Source} $\xv_A$ & {\bf Target}  $\xv_B$
\tabularnewline
a girl makes a silly face & two soccer players are playing soccer 
\tabularnewline
 \midrule
\tabularnewline
{\bf Input} $\xv_C$  & {\bf Output} $\xv_D$  \tabularnewline
$\bullet$ a girl poses for a picture &
$\bullet$ \textcolor{blue}{ two soccer players are at a soccer game.} 
\tabularnewline
$\bullet$  a girl in a blue shirt is taking pictures of a microscope & 
$\bullet$  \textcolor{blue}{ two football players in blue uniforms are at a field hockey game} \tabularnewline
$\bullet$ a woman with a red scarf looks at the stars & 
$\bullet$ \textcolor{blue}{two men in white uniforms are field hockey players}
\tabularnewline
$\bullet$ a boy is taking a bath & 
$\bullet$ \textcolor{blue}{ two baseball players are at the baseball diamond} \tabularnewline
$\bullet$ a little boy is eating a bowl of soup & 
$\bullet$ \textcolor{blue}{ two men are in baseball practice} 
\tabularnewline
$\bullet$ a mother is feeding her baby & 
$\bullet$ \textcolor{blue}{ football players are at home}
\tabularnewline
$\bullet$  a black dog is running across a field in the middle of a snowy field  & 
$\bullet$ \textcolor{blue}{ two white and black soccer players are  in the field in a soccer field} \tabularnewline
$\bullet$ some dogs are traveling to their owners & 
$\bullet$ \textcolor{blue}{dogs are in the field playing baseball} 
\tabularnewline
$\bullet$ the men were sitting on the bench at the gym for a long time & 
$\bullet$ \textcolor{blue}{ men on the field are playing in the league championship game}
\end{adjustbox}
\end{tcolorbox}
	\caption{Sentence transfer via arithmetic operation in the latent space. The output sentences are in \textcolor{blue}{blue}. In this example, we see two type of style transition: (1) from singular to plural subject, and (2) from daily-life activity to sport. }
	\label{table:transfer_example2}
\end{minipage}
\vspace{-0mm}
\end{table*}

\begin{table*}[t!]\centering
\begin{minipage}{16cm}\vspace{0mm}    \centering
\begin{tcolorbox} 
\vspace{-2mm}
\begin{adjustbox}{scale=0.8,tabular=p{8cm} p{10cm},center}
{\bf Source} $\xv_A$ & {\bf Target}  $\xv_B$
\tabularnewline
people are walking near a road. & a girl is riding a small white horse in a park with a large group of people
\tabularnewline
 \midrule
\tabularnewline
{\bf Input} $\xv_C$  & {\bf Output} $\xv_D$  \tabularnewline
$\bullet$ some people are holding cameras &
$\bullet$ \textcolor{blue}{ a girl in a black and white costume is performing a trick on a toy gun.} 
\tabularnewline
$\bullet$ people are attending church & 
$\bullet$  \textcolor{blue}{ a young girl is participating in a martial arts competition in the middle of the night.} \tabularnewline
$\bullet$ people eat at a restaurant. & 
$\bullet$ \textcolor{blue}{ a girl plays a \texttt{[UNK]} in a carnival in a city.}
\tabularnewline
$\bullet$ the dancers are asleep & 
$\bullet$ \textcolor{blue}{ the female ballet dancer is performing a ballet in the middle of a ballet class.}
\tabularnewline
$\bullet$ two dogs are reunited & 
$\bullet$ \textcolor{blue}{ a young girl is the first to capture a black and white dog in a black and white toy.} \tabularnewline
$\bullet$ a person is fishing for water. & 
$\bullet$ \textcolor{blue}{ a girl is flying a kite into a tropical storm with a tropical storm.} 
\tabularnewline
$\bullet$ a mother and daughter laugh as they walk home & 
$\bullet$ \textcolor{blue}{ a young blond-haired girl is rescued from a sad death by a young blond-haired girl in a karate ballet costume.} \tabularnewline
$\bullet$ a female gymnast is performing for a crowd & 
$\bullet$ \textcolor{blue}{ a young girl is a solo performer in a karate ballet ballet performance in a ballet performance} 
\tabularnewline
$\bullet$ a small dog is in water & 
$\bullet$ \textcolor{blue}{ a little girl is a golden retriever in a blue and white striped striped swimsuit}
\end{adjustbox}
\end{tcolorbox}
	\caption{Sentence transfer via arithmetic operation in the latent space. The output sentences are in \textcolor{blue}{blue}. In this example, we see two type of style transition: (1) from  plural/old to singular/young subject, or  and (2) sentences are expended. }
	\label{table:transfer_example3}
\end{minipage}
\vspace{-0mm}
\end{table*}

\begin{table*}[t] 
    \begin{tcolorbox}[fontupper=\footnotesize, fontlower=\footnotesize]
      \hspace{-7mm}
    \begin{tabular}{l l}
~ & {\bf Positive} \tabularnewline
~ & our favorite place to get great coffee and taterts.
\tabularnewline
~ & the best brunch you will find in vegas.
\tabularnewline
~ & the best breakfast with meats is awesome!
\tabularnewline
~ & great samosas and serve as a regular!
\tabularnewline
~ & great place to meet up with a custom bean \& wine.
\tabularnewline
~ & a great selection of chinese food and always happy.
\tabularnewline
~ & the free wi-fi is amazing as well!
\tabularnewline
~ & great staff and freshly made latte is a must.
\tabularnewline
~& love the fresh staff as well!
\tabularnewline
~ & highly recommend the place and sunbeams!
\tabularnewline
~ & the staff is always great with homemade paesadillas.
	\end{tabular}
	\end{tcolorbox}
	%
    \begin{tcolorbox}[fontupper=\footnotesize, fontlower=\footnotesize]
      \hspace{-7mm}
    \begin{tabular}{l l}
~ & {\bf Negative} \tabularnewline
~ & not only did you get a headache upstairs, they were disgusting.
\tabularnewline
~ & once i realized the pizza wasn't decent, i cancelled.
\tabularnewline
~ & instead of going to the bathroom you couldn't find anything.
\tabularnewline
~ & tonight i was unable to give the pizza any less.
\tabularnewline
~ & i didn't even bother to find a $\$$ [num] frozen pizza.
\tabularnewline
~ & no wonder i was dropped off at laundry.
\tabularnewline
~ & not only was this place freezing, but the salad sucked.
\tabularnewline
~ & then [num] bucks was ruined in my mouth.
\tabularnewline
~& love the fresh staff as well!
\tabularnewline
~ & once you asked for chipotle its out of control.
\tabularnewline
~ & another thing i refused to eat.
	\end{tabular}
	\end{tcolorbox}
	\caption{Label-conditional text generation on Yelp dataset. The top block shows the positive reviews, and bottom block shows the negative reviews. }
	\label{table:label_conditional_example1}
	\vspace{2mm}
\end{table*}



\end{document}